\documentclass[11pt]{article}

\usepackage[preprint]{acl}

\usepackage{times}
\usepackage{latexsym}

\usepackage[T1]{fontenc}

\usepackage[utf8]{inputenc}

\usepackage{microtype}

\usepackage{inconsolata}

\usepackage{amsmath} 
\usepackage{amssymb}
\usepackage{algorithm}
\usepackage{algpseudocode}
\usepackage{float}    
\usepackage{wrapfig}  

\usepackage{graphicx}
\usepackage{xcolor}         
\usepackage{booktabs} 
\usepackage{multirow}       
\usepackage{cleveref}
\usepackage{comment}

\usepackage[most]{tcolorbox}

\usepackage{tabularx}
\usepackage{makecell}

\usepackage[table]{xcolor}

\definecolor{best}{rgb}{0.96, 0.57, 0.58}
\definecolor{second}{rgb}{0.98, 0.78, 0.57}
\definecolor{third}{rgb}{1.0, 1.0, 0.56}

\usepackage{makecell}
\definecolor{gainstrong}{HTML}{81C784}
\definecolor{gainmid}{HTML}{C8E6C9}
\definecolor{gainmild}{HTML}{E8F5E9}
\definecolor{lossmild}{HTML}{FFEBEE}
\definecolor{lossmid}{HTML}{FFCDD2}
\definecolor{lossstrong}{HTML}{EF9A9A}

\newcommand{\modelicon}[1]{%
    \raisebox{-0.15em}{
        \includegraphics[height=1em]{figs/icons/#1}%
    }\hspace{0.4em}
}

\definecolor{ogaCapStrong}{HTML}{2CA02C} 
\definecolor{ogaCapMod}{HTML}{F08C1A}    
\definecolor{ogaCapWeak}{HTML}{D62728}   
\definecolor{ogaCapNone}{HTML}{BFBFBF}   

\title{OmniGameArena: A Unified UE5 Benchmark for VLM Game Agents with Improvement Dynamics}

\author{
  \textbf{Mingxian Lin\textsuperscript{1}},
  \textbf{Shengju Qian\textsuperscript{2,}\,$^\ddagger$},
  \textbf{Yuqi Liu\textsuperscript{3}},
  \textbf{Yi-Hua Huang\textsuperscript{1}},
  \textbf{Yiyu Wang\textsuperscript{2}},
  \textbf{Wei Huang\textsuperscript{1}},
\\
  \textbf{Yitang Li\textsuperscript{4}},
  \textbf{Fan Zhang\textsuperscript{3}},
  \textbf{Zeyu Hu\textsuperscript{2}},
  \textbf{Lingting Zhu\textsuperscript{2}},
  \textbf{Xin Wang\textsuperscript{2}},
  \textbf{Xiaojuan Qi\textsuperscript{1,}\,$^\dagger$}
\\
\\
  \textsuperscript{1}The University of Hong Kong,
  \textsuperscript{2}LIGHTSPEED,
\\
  \textsuperscript{3}The Chinese University of Hong Kong,
  \textsuperscript{4}Tsinghua University
\\[6pt]
  \small{$\ddagger$\,Project Leader \qquad $\dagger$\,Corresponding Author}
\\[6pt]
  \small{Project Page: \url{https://mxlin043.github.io/OmniGameArena/}}
  \\[10pt]
}

\begin{document}
\maketitle
\vspace{6pt}    %

\begin{abstract}
Vision-language model (VLM) agents are increasingly deployed in
interactive game environments. Yet game benchmarks for VLM agents
typically report a single first-attempt score per (agent, game)
pair, focus on single-agent Solo play, and lack unified protocols
for evaluating heterogeneous agent classes (commercial VLMs,
open-weight VLMs, and specialized game policies) on the same
footing. We address these gaps with OmniGameArena, a
real-time benchmark of twelve newly built Unreal Engine~5 games
spanning Solo (7), PvP (3), and Coop (2) with unified action
interfaces, and the Improvement Dynamics Curve (IDC),
an agentic-reflection harness in which a tool-using reflector
LLM autonomously refines a bounded skill prompt across multiple
rounds. Beyond cold-start leaderboard scores, IDC exposes two
additional observables for each (agent, game) pair: how the
score evolves across reflection rounds, and how the learned
skill behaves on held-out task variants. We report these
observables for twelve VLM agents on the cold-start leaderboard and
four top agents under IDC.
\end{abstract}

\begin{figure*}[t]
\centering    \includegraphics[width=1.0\textwidth]{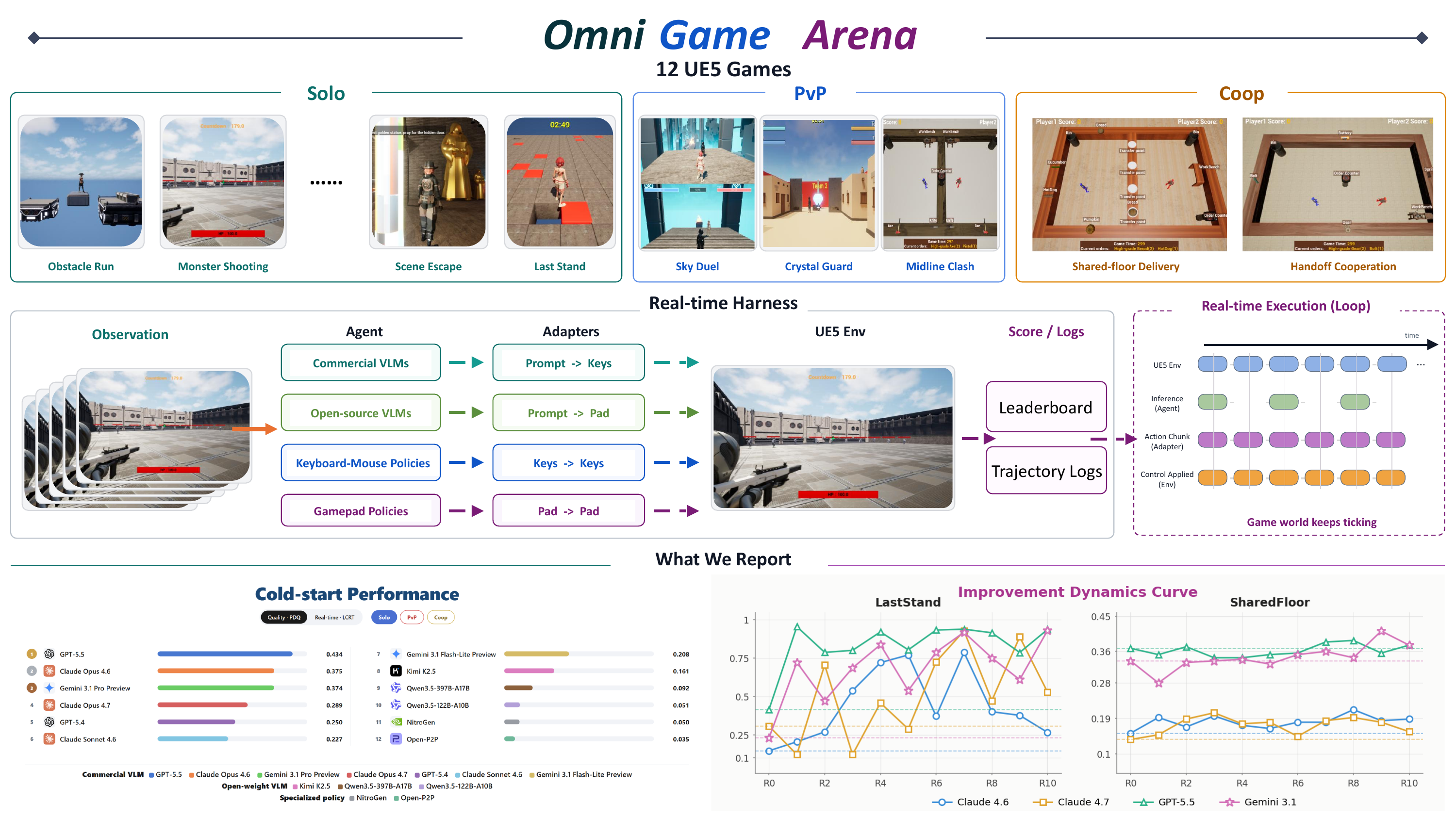}
\caption{OmniGameArena at a glance. Twelve newly built UE5 games
span Solo (7), PvP (3), and Coop (2) regimes (top).
Heterogeneous agents (commercial VLMs, open-weight VLMs,
keyboard-mouse policies, and gamepad policies) connect to the
same real-time UE5 environment through documented adapters
(middle). Evaluation reports the cold-start leaderboard and the
Improvement Dynamics Curve (IDC) under multi-round reflection
(bottom).}
\label{fig:teaser}
\vspace{-3mm}
\end{figure*}

\section{Introduction}
\label{sec:introduction}

Foundation models are increasingly evaluated by how they act,
not only by what they answer, and games are a natural stress
test for this shift~\citep{wang2023voyager,tan2024cradle,paglieri2024balrog}:
an agent must read a changing visual scene, choose actions under
time pressure, plan across delayed rewards, and adapt when the
environment resists. Game benchmarks now span text-only worlds,
2D grid suites, and 3D open environments built on
existing commercial titles, and have driven rapid progress in
vision-language game agents~\citep{tan2025lumine,magne2026nitrogen,wang2025game}.

Yet current benchmarks rarely measure two properties that matter
for deploying these agents. Most report a single first-attempt
score per (agent, game) pair, leaving invisible the trajectory
by which an agent improves under repeated interaction with the
same task. They also lean heavily toward single-agent Solo play,
while adversarial (PvP) and cooperative (Coop) regimes remain
underrepresented even though they probe distinct capabilities
such as opponent modeling, role assignment, and recovery from a
teammate's mistakes. Whether an agent can adapt under repeated
reflection, and whether it can do so in adversarial or
cooperative settings, therefore remains largely unmeasured.

We address both with \textbf{OmniGameArena}, a real-time
benchmark of twelve newly built Unreal Engine~5 games spanning
Solo, PvP and Coop, and the \textbf{Improvement
Dynamics Curve (IDC)}, an agentic-reflection harness built on
top of it. The twelve games are authored for this benchmark
rather than reused from public titles, lowering the risk of
pre-training leakage, and share unified action interfaces
(keyboard-mouse, gamepad) so that commercial
VLMs, open-weight VLMs, and specialized game policies can all
be evaluated under matched environment conditions. The IDC
harness runs each (agent, game) instance for multiple rounds:
the agent plays $K$ episodes under a current skill prompt,
after which a reflector LLM inspects the trajectories through
tool-use, deciding on its own what to read and when to stop,
before refining the skill for the next round. We report both
the per-round score sequence (the IDC of that instance) and a
transfer score on held-out task variants.

Across twelve agents on the cold-start leaderboard, no single
VLM dominates, and commercial agents hold a wide gap over
open-weight VLMs and specialized policies. Among the four top
agents that we run through IDC, all four improve over their
cold-start baseline through reflection, yet peak performance is
typically reached mid-curve rather than at the final round.
Most notably, origin-task improvement and held-out variant
transfer can diverge in our experiments; this divergence is
hidden by single-round leaderboard scores and is a central
observable IDC exposes.

To summarize, our contributions are threefold:
(i) \textbf{OmniGameArena}, a twelve-game UE5 benchmark
spanning Solo, PvP, and Coop with unified action interfaces and
game instances built specifically for this benchmark;
(ii) the \textbf{IDC harness}, an agentic-reflection framework
whose autonomous tool-use reflector refines a bounded skill
prompt across $R$ rounds, with persistent memory and best-skill
rollback; and (iii) an empirical study across twelve agents
showing that leadership rotates across games and that
origin-task gain does not by itself predict held-out variant
transfer.

\begin{figure*}[t]
    \centering
    \includegraphics[width=\textwidth]{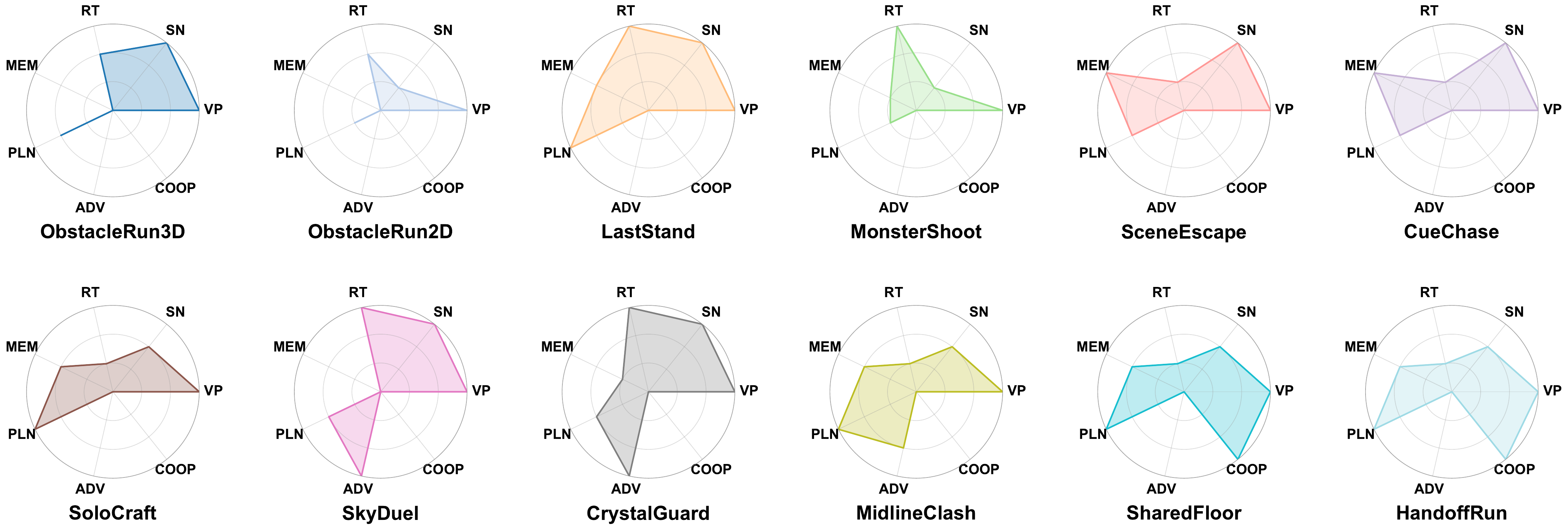}
    \caption{Radar charts of the 12 OmniGameArena games across seven capability dimensions. The abbreviations are: VP = Visual Perception, SN = Spatial Navigation, RT = Reaction, MEM = Memory, PLN = Planning, ADV = Adversarial, and COOP = Cooperation. Each dimension is scored from 0 to 3.}
    \label{fig:omni_game_arena_radar}
\end{figure*}

\section{Related Work}
\label{sec:related_work}

\noindent\textbf{Benchmarks in game environments.}
Interactive games have served as AI testbeds since the rise of
reinforcement learning and now anchor evaluations of Large
Language Models (LLMs) and Vision-Language Models (VLMs). Early
LLM evaluations were text-only
\citep{huang2024far,wu2023smartplay,hu2024gamearena}, effective
for logical reasoning but lacking visual grounding. 2D-grid
suites such as BALROG \citep{paglieri2024balrog} and
LVLM-Playground \citep{wang2025large} added spatial and
multimodal demands, and more recent suites built on Minecraft or
general visual benchmarks (V-MAGE \citep{zheng2025v}, Cradle
\citep{tan2024cradle}, VideoGameBench
\citep{zhang2025videogamebench}) extended evaluation into 3D
open worlds with long-horizon planning from pixels. 
Beyond game playing, related benchmarks probe embodied reasoning and action in complex 3D environments \citep{lin2025embrace,zhu2026assetformer} and unified reasoning across video-generation models \citep{luo2025v}, but neither targets multi-regime, real-time game interaction.
Two limitations of these benchmarks motivate our work: most reuse
existing commercial titles, leaving them exposed to pre-training
contamination; and few cover Solo, PvP, and Coop regimes in a
single real-time environment. OmniGameArena addresses both with
twelve newly built UE5 games that span all three interaction
regimes.

\noindent\textbf{Game-playing LLM and VLM agents.}
Early LLM game agents operated in text-only environments
\citep{hausknecht2020interactive,tsai2023can} and 2D grid worlds
\citep{feng2023chessgpt,kuttler2020nethack}. Voyager
\citep{wang2023voyager} and MineDojo \citep{fan2022minedojo}
extended LLM agents to 3D Minecraft, but the heavy per-game
engineering they require limits cross-game generality. The
current VLM-agent generation
\citep{li2025jarvis,bai2026webgym,wang2025game,tan2025lumine,magne2026nitrogen}
controls GUI or keyboard-mouse interfaces directly across diverse
3D worlds: Game-TARS \citep{wang2025game} is pre-trained on over
500B tokens of multimodal gameplay; NitroGen
\citep{magne2026nitrogen} on $40{,}000$ hours of gameplay video
across $1{,}000+$ titles; and Lumine \citep{tan2025lumine}
executes hours-long real-time missions in 3D environments. These
agents motivate, but also outpace, the standardized cross-game
evaluation infrastructure we provide.

\noindent\textbf{Reflection and Self-improvement.}
Most game-agent benchmarks report a single-shot score, obscuring
whether and how fast an agent improves with repeated interaction.
Reflection-based methods address this without weight updates by
accumulating natural-language summaries of past experience.
Reflexion \citep{shinn2023reflexion} converts episode feedback
into verbal self-critiques; Self-Refine
\citep{madaan2023selfrefine} iteratively rewrites model outputs;
ExpeL \citep{zhao2024expel} extracts task-level insights;
Voyager's skill library \citep{wang2023voyager} accumulates
reusable code skills inside Minecraft; and GameVerse
\citep{zhang2026gameverse} reports a single with-vs-without
reflection comparison per game. Our work aligns with the recently
articulated \emph{heuristic learning} paradigm
\citep{weng2026learning_beyond_gradients}, which views LLM-driven self-improvement
as a learning process operating on explicit artifacts (prompts,
code, memory) rather than weights. Our IDC harness instantiates
this paradigm in three concrete ways: (i) reflection runs for
multiple rounds, producing a full score trajectory rather than a
before-after comparison; (ii) the reflector is itself an LLM with
tool-use that decides what to inspect rather than executing a
fixed-template script; and (iii) we test the resulting skill on
held-out task variants per game, revealing differences in skill
style that single-number metrics hide.

\begin{table*}[t]
\centering
\footnotesize
\renewcommand{\arraystretch}{1.1}
\begin{tabular}{lp{7.cm}p{6cm}}
\toprule
\textbf{Game} & \textbf{Description} & \textbf{Evaluation} \\ 
\midrule

\rowcolor{gray!15} 
\multicolumn{3}{l}{\textit{Solo}} \\
ObstacleRun3D & A 3D parkour game where the agent navigates to a finish line while avoiding physical obstacles. & $\frac{x_a - x_{start}}{x_{finish} - x_{start}}$, where $x_a$: agent pos., $x_{start}$: start, $x_{finish}$: target \\
ObstacleRun2D & A 2D side-scrolling platformer where the agent must reach the end of a linear level. & $\frac{x_a - x_{start}}{x_{finish} - x_{start}}$, where $x_a$: agent pos., $x_{start}$: start, $x_{finish}$: target \\
LastStand & A platform survival game where the agent must avoid hazards and falling off. & $\frac{t_{survive}}{T_{max}}$, where $t_{survive}$: time survived, $T_{max}$: max duration \\
MonsterShoot & A survival shooting game to locate and eliminate hostile entities while avoiding damage. & $\frac{D_e}{H_{total}}$, where $D_e$: effective damage, $H_{total}$: total enemy health \\
SceneEscape & A scene-based puzzle game requiring the completion of NPC-assigned tasks to escape. & $\frac{n}{N}$, where $n$: completed tasks, $N$: total tasks \\
CueChase & A third-person exploration game to locate and activate hidden triggers across the map. & $\frac{k}{K}$, where $k$: activated triggers, $K$: total triggers \\
SoloCraft & A logistics game where the agent collects, prepares, and delivers items to fulfill orders. & $\frac{v_{delivered}}{V_{target}}$, where $v_{delivered}$: fulfilled value, $V_{target}$: target value \\ 
\midrule

\rowcolor{gray!15} 
\multicolumn{3}{l}{\textit{PvP}} \\
SkyDuel & A direct 1v1 combat game where the agent must engage and defeat an opponent. & $\frac{h_{self}}{H_{max}}$, where $h_{self}$: agent remaining health, $H_{max}$: agent max health \\
CrystalGuard & A symmetric attack-and-defend game to destroy the opponent's crystal core while protecting one's own. & $\frac{h^{own}_{core}}{H_{max}}$, where $h^{own}_{core}$: own core health, $H_{max}$: core max health \\
MidlineClash & A competitive logistics game where two agents race to fulfill resource orders in a shared environment. & $\frac{s_a}{S_{target}}$, where $s_a$: agent score, $S_{target}$: target score \\ 
\midrule

\rowcolor{gray!15} 
\multicolumn{3}{l}{\textit{Cooperation}} \\
SharedFloor & A symmetric cooperative game where agents share a space and capabilities to fulfill delivery orders. & $\frac{v_{delivered}}{V_{team}}$, where $v_{delivered}$: fulfilled value, $V_{team}$: team target value \\
HandoffRun & An asymmetric cooperative game requiring agents to pass items across restricted areas based on distinct roles. & $\frac{v_{delivered}}{V_{team}}$, where $v_{delivered}$: fulfilled value, $V_{team}$: team target value \\ 

\bottomrule
\end{tabular}
\vspace{-2mm}
\caption{Summary of 12 Interactive Games.}
\label{tab:games_summary}
\vspace{-2mm}
\end{table*}

\section{OmniGameArena}
\label{sec:omnigamearena}

We introduce OmniGameArena, a suite of twelve custom Unreal Engine (UE5) games spanning Solo, PvP, and Coop regimes (\Cref{sec:oga:games}) to systematically evaluate distinct capability axes of vision-based game agents using robust, continuous progress metrics. 
Furthermore, to address the critical challenge of evaluation contamination, we detail proactive data avoidance strategies and rigorous empirical analysis (\Cref{sec:oga:contam}) to ensure the integrity and novelty of our benchmark.

\subsection{Game Suite and Evaluation Metrics}
\label{sec:oga:games}

The OmniGameArena suite comprises twelve visually rich and physically complex environments developed in Unreal Engine 5. Each game is purposefully designed to isolate and test specific subsets of embodied capabilities, ranging from solitary spatial reasoning to complex multi-agent cooperation, as shown in \Cref{fig:omni_game_arena_radar}.  
Game progress is uniformly normalized to a continuous scale of $[0, 1]$, providing a consistent metric across highly diverse tasks. We provide a brief description and its evaluation protocol for each game, as demonstrated in \Cref{tab:games_summary}.

\subsection{Contamination Avoidance}
\label{sec:oga:contam}

We adopt a proactive approach to mitigate data contamination during the benchmark design phase. We begin by conducting a \emph{web-exposure audit} to search for exact game names, task phrases, rule descriptions, and scoring events prior to release, ensuring these specific elements are strictly excluded from the benchmark. To guarantee environment novelty, we construct entirely new games using Unreal Engine 5 (UE5). For visual content, we utilize a mixture of bespoke and off-the-shelf UE5 marketplace assets. 
However,
the combination of assets, the design of the level geometry, the execution order of scripts, and the criteria for success are uniquely designed, ensuring the evaluation scenarios cannot be memorized from pre-training data.

\begin{table}[t]
\centering
\footnotesize
\setlength{\tabcolsep}{2pt}
\begin{tabular}{lccc}
\toprule
\textbf{Benchmark} & \textbf{Games(\#)} & \textbf{Recog.(\%)$\downarrow$} & \textbf{Mech. (\%)$\downarrow$}  \\
\midrule
BALROG        & \hphantom{0}6  & \hphantom{0}66.7    & 100.0 \\
LMGame-Bench  & \hphantom{0}6  & 100.0               & 100.0 \\
ORAK          & 12             & 100.0               & 100.0 \\
OmniGameArena & 12             & \hphantom{00}0.0    & \hphantom{0}50.0 \\
\bottomrule
\end{tabular}
\vspace{-2mm}
\caption{Contamination analysis when given screenshots. Recog.: Percentage of games recognized; Mech.: Percentage of games where underlying mechanics were successfully described.}
\label{tab:contamination}
\vspace{-4mm}
\end{table}

\textbf{Contamination Analysis.}
\label{sec:oga:contam_analysis}
To empirically verify the effectiveness of our avoidance strategies, we conduct a contamination analysis focusing on visual novelty and rule leakage. First, we provide a representative model (e.g., Gemini) with screenshots of games to assess whether it can recognize the game name, confirming the visual novelty of the tasks. Second, we evaluate whether the model can successfully describe the underlying mechanics of the games based purely on visual inputs, which tests for the leakage of memorized game rules. 
As shown in Table~\ref{tab:contamination}, we compare existing benchmarks \citep{paglieri2024balrog,park2025orak,hu2025lmgamebench} against our proposed OmniGameArena across these two tests. The
results clearly demonstrate that the games within the existing benchmarks are highly recognizable, allowing the model to
retrieve their mechanics directly from memory. In contrast, OmniGameArena exhibits a $0.0\%$ recognition rate and significantly
reduced mechanics leakage ($50.0\%$). These results suggest that OmniGameArena substantially mitigates the risk of pre-training
contamination, reducing the influence of memorized priors on agent evaluation.


\begin{figure*}[t]
    \centering
    \includegraphics[width=1.0\textwidth]{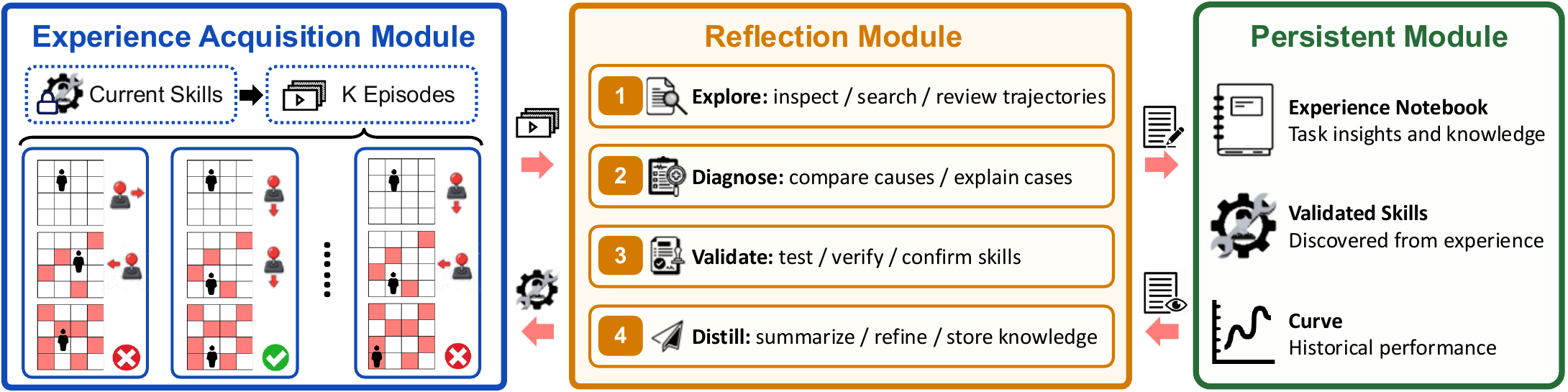}
        \caption{
            Overview of the Improvement Dynamics Curve (IDC) harness. The
            \emph{experience acquisition} module (left) runs $K$ episodes
            under the current skill. The \emph{reflection} module (center)
            reads both the new trajectories and the persistent state
            (notebook and prior skill), then runs four autonomous stages
            (Explore, Diagnose, Validate, Distill) to produce a refined
            skill. The \emph{persistent} module (right) stores the
            experience notebook, validated skills, and per-round score
            curve, which seed the next round.
            }
    \label{fig:self_evolving_agent}
\end{figure*}

\section{Game Agent Harness}
\label{sec:harness}

OmniGameArena specifies the games; the harness specifies how an
agent plays them. The harness has two layers: a per-episode loop
(\S\ref{sec:harness:episode}) that drives any agent during
cold-start runs, and a reflective outer loop
(\S\ref{sec:harness:idc}) whose round-level scores form the
Improvement Dynamics Curve studied in
\S\ref{sec:experiments:idc}.

\subsection{Per-Episode Loop}
\label{sec:harness:episode}

OmniGameArena exposes a Gym-like interface. At step $t$ the
harness receives observation
$o_t = (I_t, w_t, h_t, \tau_t)$, where $I_t$ is the RGB frame at
resolution $w_t \times h_t$ and $\tau_t$ is its capture
timestamp. The agent emits an action $a_t$ (a chunked
keyboard-mouse action for VLMs), which
the engine executes before returning $o_{t+1}$; the loop
terminates on $\texttt{done}$.

\paragraph{Bounded visual-action history.} For VLM agents, the
harness maintains a sliding window of the last $L$
observation-response pairs:
\[
\mathcal{H}_t = [(I_{t-L}, y_{t-L}), \ldots, (I_{t-1}, y_{t-1})],
\]
where $y_i$ is the raw VLM response at step $i$. At step $t$, the
VLM is prompted with system instructions, an optional skill
prompt $m$, the history $\mathcal{H}_t$, and the current frame
$I_t$, for up to $L+1$ images per call. The current frame is
appended to history only after $y_t$ is returned.

\subsection{Improvement Dynamics Curve}
\label{sec:harness:idc}

IDC wraps the per-episode loop with a reflective outer loop
organized into three modules
(Figure~\ref{fig:self_evolving_agent}): an \emph{experience
acquisition} module that runs $K$ episodes under the current
skill, a \emph{reflection} module that converts the resulting
trajectories into a refined skill, and a \emph{persistent}
module that carries state across rounds. We deliberately keep
the loop lightweight yet functionally complete: a small fixed
tool surface gives the reflector full agentic autonomy without
prescribing an inspection script.

Let $\pi_\theta$ denote a VLM with frozen weights $\theta$. At
round $r$ the agent is conditioned on a skill prompt $m_r$ and
acts according to
$a_t \sim \pi_\theta(a_t \mid o_t, \mathcal{H}_t, m_r)$; the
skill $m_r$ is fixed throughout the round.

\noindent\textbf{Experience acquisition.}
Each round contains $K$ episodes, yielding trajectories
$\tau_r = \{\tau_{r,1}, \ldots, \tau_{r,K}\}$ and round score
\[
S_r = \frac{1}{K} \sum_{k=1}^{K} s(\tau_{r,k}).
\]
Round $r=0$ uses an empty skill ($m_0 = \emptyset$) and serves
as the cold-start baseline.

\begin{table*}[t]
\centering
\small

\setlength{\tabcolsep}{4pt}
\renewcommand{\arraystretch}{1.1}
\resizebox{\textwidth}{!}{%
\begin{tabular}{lccccccc}
\toprule
Agent & ObstacleRun2D & ObstacleRun3D & LastStand & MonsterShoot & SceneEscape & CueChase & SoloCraft \\
\midrule
\modelicon{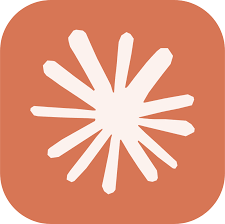}~Claude Opus 4.7 \citep{anthropic2026opus47}                 
& {$0.220_{\pm 0.218}$} & {$0.094_{\pm 0.040}$} & \cellcolor{second}{$0.308_{\pm 0.146}$} & \cellcolor{third}{$0.400_{\pm 0.087}$} & {$0.460_{\pm 0.152}$} & {$0.380_{\pm 0.192}$} & \cellcolor{third}{$0.160_{\pm 0.028}$} \\
\modelicon{claude}~Claude Opus 4.6   \citep{anthropic2026opus46}               
& \cellcolor{second}{$0.338_{\pm 0.003}$} & \cellcolor{second}{$0.172_{\pm 0.094}$} & {$0.147_{\pm 0.044}$} & {$0.362_{\pm 0.079}$} & {$0.540_{\pm 0.134}$} & \cellcolor{best}{$0.840_{\pm 0.102}$} & \cellcolor{second}{$0.228_{\pm 0.023}$} \\
\modelicon{claude}~Claude Sonnet 4.6     \citep{anthropic2026sonnet46}           
& {$0.075_{\pm 0.002}$}  & \cellcolor{best}{$0.200_{\pm 0.138}$} & {$0.144_{\pm 0.050}$} & {$0.166_{\pm 0.048}$} & {$0.440_{\pm 0.207}$} & {$0.440_{\pm 0.215}$} & {$0.124_{\pm 0.033}$} \\
\modelicon{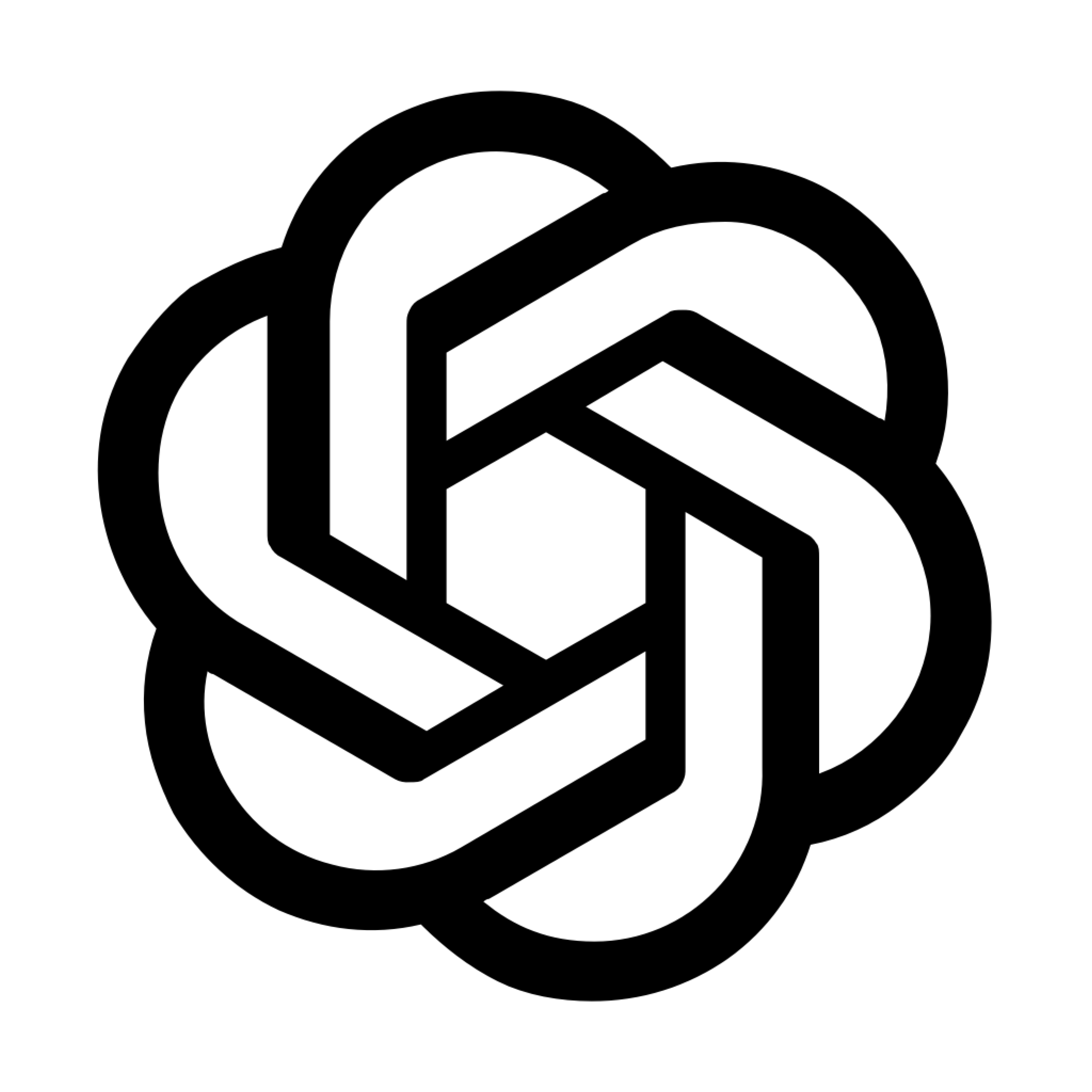}~GPT-5.5         \citep{openai2026gpt55}                 
& \cellcolor{best}{$0.473_{\pm 0.121}$} & {$0.133_{\pm 0.007}$} & \cellcolor{best}{$0.416_{\pm 0.257}$}  & \cellcolor{second}{$0.464_{\pm 0.064}$} & \cellcolor{best}{$0.720_{\pm 0.370}$} & \cellcolor{third}{$0.580_{\pm 0.098}$} & \cellcolor{best}{$0.252_{\pm 0.023}$} \\
\modelicon{openai}~GPT-5.4          \citep{openai2026gpt54}                  
& {$0.122_{\pm 0.070}$} & {$0.089_{\pm 0.037}$} & {$0.148_{\pm 0.070}$} & {$0.326_{\pm 0.090}$} & \cellcolor{second}{$0.680_{\pm 0.045}$} & {$0.300_{\pm 0.063}$} & {$0.084_{\pm 0.017}$} \\
\modelicon{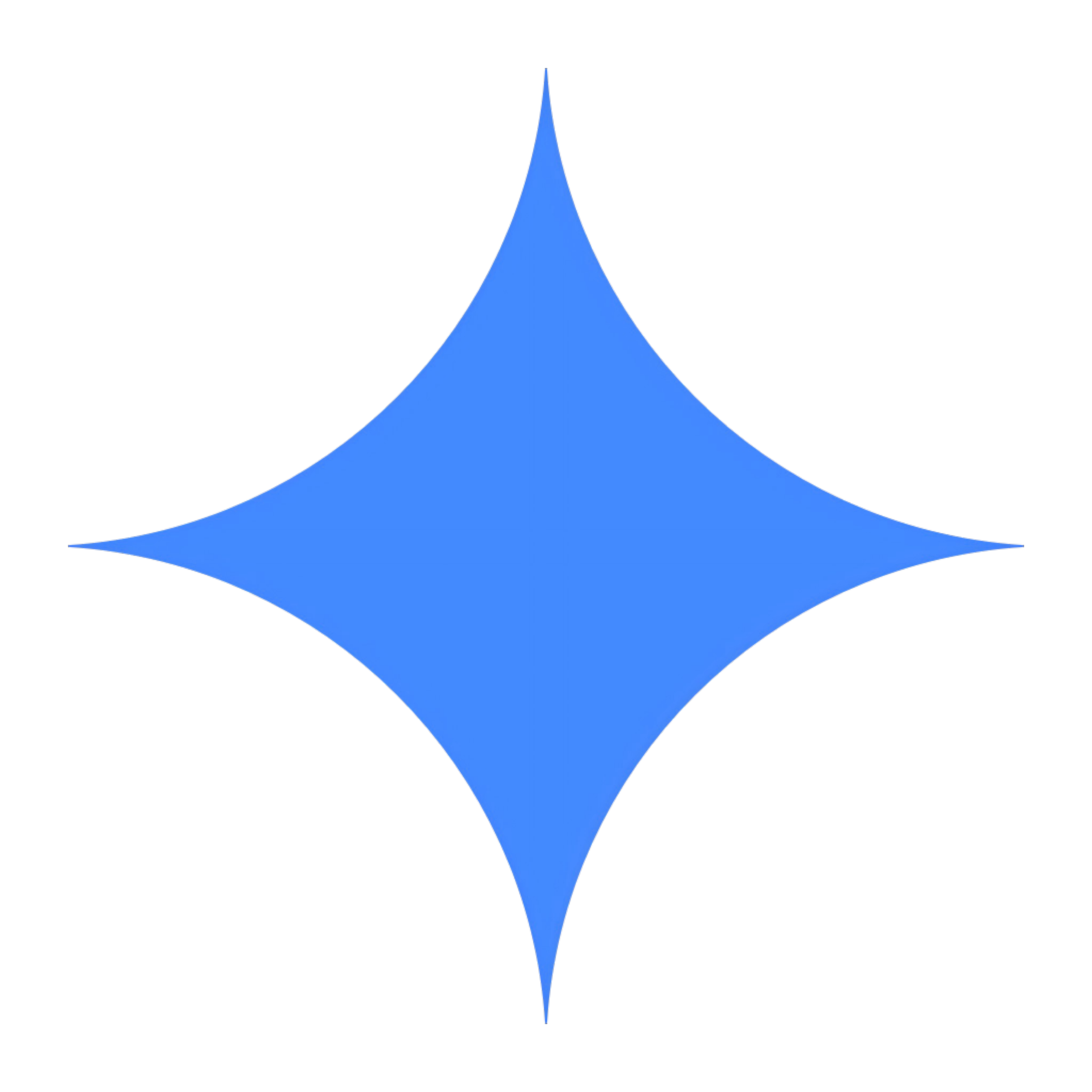}~Gemini 3.1 Pro Preview  \citep{google2026gemini31pro}          
& {$0.102_{\pm 0.059}$} & \cellcolor{third}{$0.165_{\pm 0.112}$} & {$0.230_{\pm 0.090}$}  & \cellcolor{best}{$0.710_{\pm 0.138}$} & \cellcolor{third}{$0.660_{\pm 0.336}$} & \cellcolor{second}{$0.600_{\pm 0.322}$} & {$0.148_{\pm 0.100}$} \\
\modelicon{gemini}~Gemini 3.1 Flash-Lite Preview \citep{google2026gemini31pro}   
& \cellcolor{third}{$0.278_{\pm 0.086}$} & {$0.097_{\pm 0.044}$} & {$0.122_{\pm 0.030}$}  & {$0.182_{\pm 0.083}$} & {$0.300_{\pm 0.158}$} & {$0.440_{\pm 0.080}$} & {$0.036_{\pm 0.022}$} \\
\modelicon{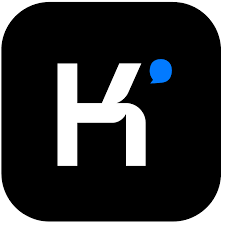}~Kimi K2.5         \citep{kimi_k25_2026}               
& {$0.109_{\pm 0.056}$} & {$0.075_{\pm 0.031}$} & \cellcolor{third}{$0.232_{\pm 0.113}$} & {$0.290_{\pm 0.075}$} & {$0.220_{\pm 0.130}$} & {$0.140_{\pm 0.102}$} & {$0.064_{\pm 0.043}$} \\
\midrule
\modelicon{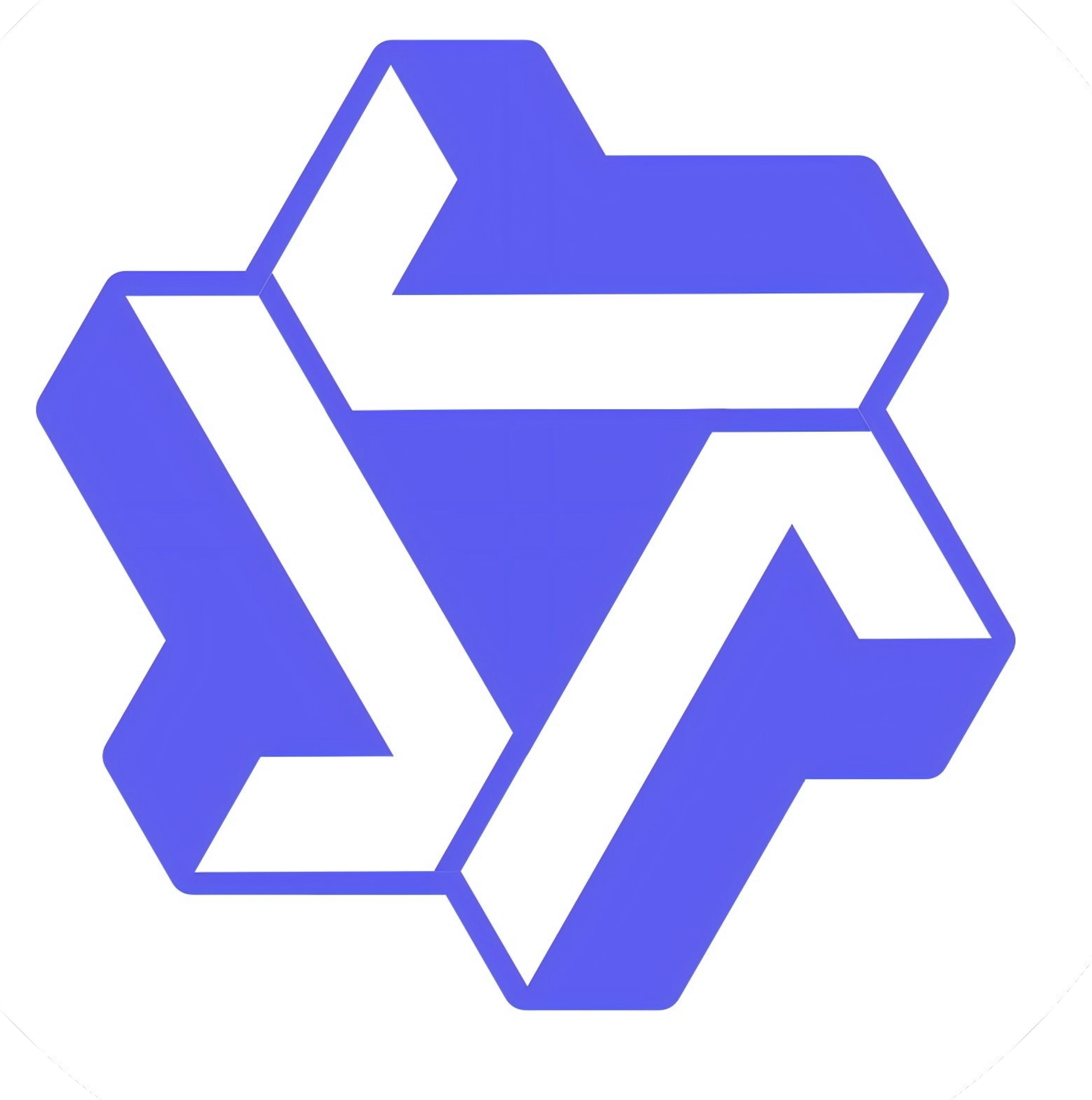}~Qwen3.5-397B-A17B          \citep{qwen2026qwen35}            
& {$0.114_{\pm 0.012}$} & {$0.112_{\pm 0.043}$} & {$0.106_{\pm 0.017}$} & {$0.072_{\pm 0.025}$} & {$0.200_{\pm 0.258}$} & {$0.040_{\pm 0.049}$} & {$0.000_{\pm 0.000}$} \\
\modelicon{qwen}~Qwen3.5-122B-A10B      \citep{qwen2026qwen35}   
& {$0.092_{\pm 0.028}$} & {$0.034_{\pm 0.032}$} & {$0.093_{\pm 0.019}$} & {$0.024_{\pm 0.019}$} & {$0.060_{\pm 0.080}$} & {$0.040_{\pm 0.049}$} & {$0.013_{\pm 0.023}$} \\
\midrule
\modelicon{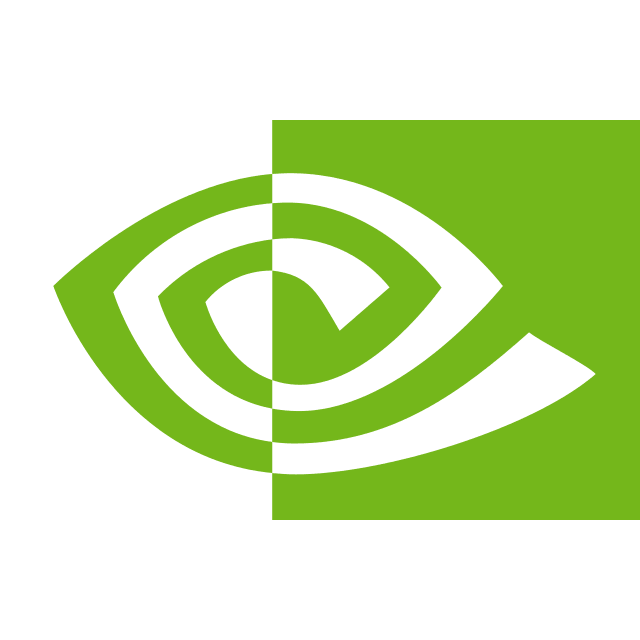}~NitroGen (gamepad)          \citep{magne2026nitrogen}
& {$0.034_{\pm 0.042}$} & {$0.065_{\pm 0.046}$} & {$0.100_{\pm 0.029}$} & {$0.014_{\pm 0.014}$} & {$0.120_{\pm 0.040}$} & {$0.020_{\pm 0.040}$} & {$0.000_{\pm 0.000}$} \\
\modelicon{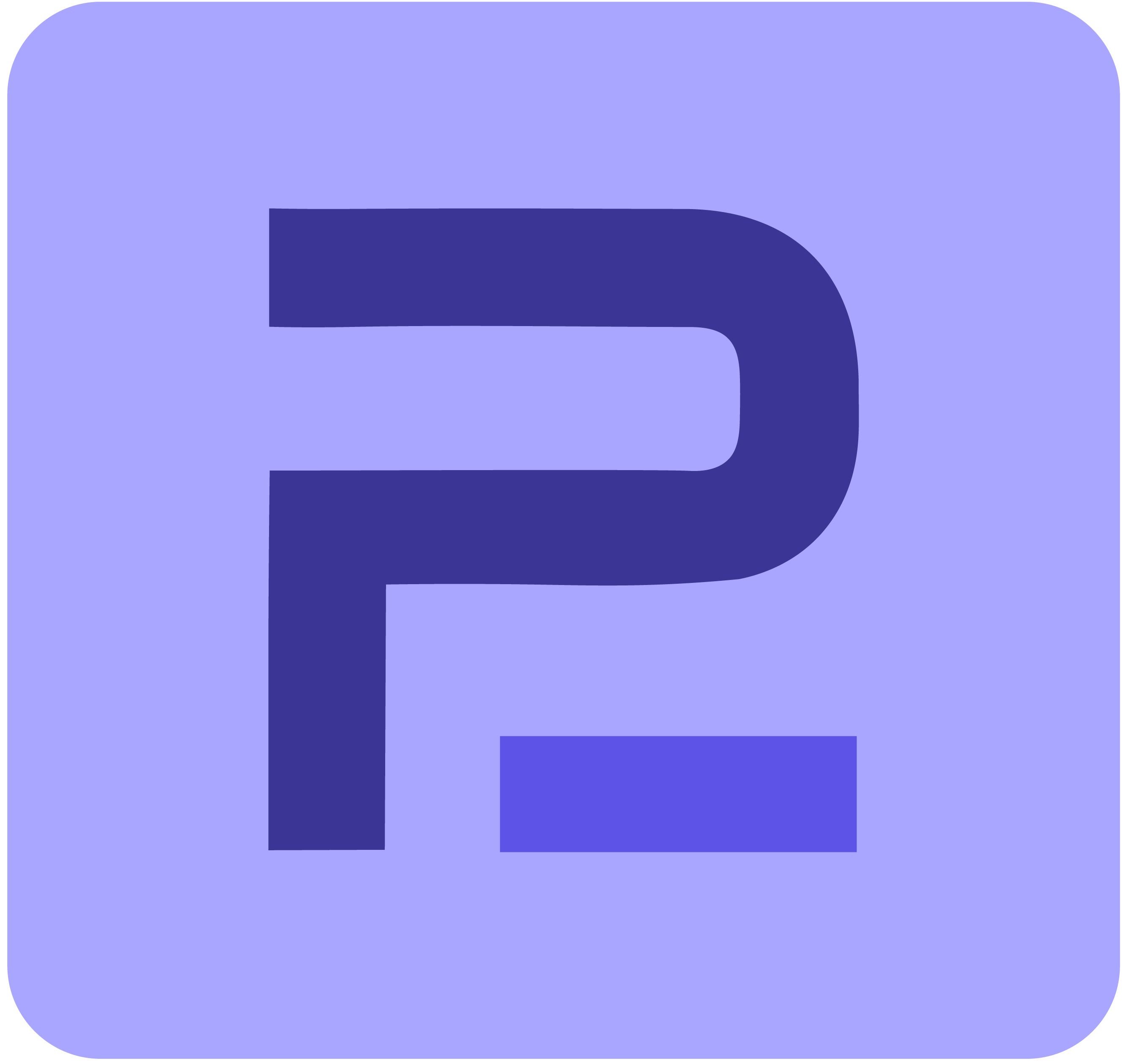}~Open-P2P (kbd-mouse)              \citep{yue2026scaling}
& {$0.056_{\pm 0.040}$} & {$0.023_{\pm 0.045}$} & {$0.063_{\pm 0.034}$} & {$0.000_{\pm 0.000}$} & {$0.100_{\pm 0.082}$} & {$0.000_{\pm 0.000}$} & {$0.000_{\pm 0.000}$} \\
\bottomrule
\end{tabular}%
}
\caption{Solo cold-start scores. Top 3 per column:
\colorbox{best}{red} $>$ \colorbox{second}{orange} $>$
\colorbox{third}{yellow}.}
\label{tab:main-solo}
\vspace{-2mm}
\end{table*}

\noindent\textbf{Reflection.}
After round $r$, the reflector $\mathcal{R}$ refines the skill
autonomously, reading the new trajectories together with the
persistent state (notebook and prior skill) and deciding for
itself which content to inspect, how many tool calls to spend,
and when to terminate. The refinement proceeds in four stages.
The \emph{Explore} stage exposes the round's per-episode trajectories through sandboxed
read-only tools (\texttt{list\_dir}, \texttt{read\_text},
\texttt{read\_image}, \texttt{grep}); the reflector chooses what
to inspect rather than executing a fixed script. The
\emph{Diagnose} stage commits an explicit list of failure modes
via \texttt{submit\_diagnosis}, separating causes from
prescriptions. The \emph{Validate} stage proposes an updated
skill and calls \texttt{validate\_skill}, an independent LLM
judge that rejects proposals which memorize map content, or contradict the diagnosis;
the reflector iterates up to five times before committing. The
\emph{Distill} stage finalizes the accepted skill $m_{r+1}$ and,
optionally, edits the notebook with durable observations. The
combination of agentic autonomy within a small, fixed tool
surface (read, diagnose, judge, write) is what keeps the
framework lightweight without sacrificing functional completeness.

\noindent\textbf{Persistent module.}
Three artifacts carry state across rounds. The \emph{experience
notebook} is a factual log written by the reflector for its own
future use (e.g., ``round $r$ episode $k$ step $t$: event''),
capped at $2000$ tokens, edited rather than appended, and never
shown to the player. The \emph{validated skills} comprise the
skill prompt $m_r$ used in the current round (player-visible,
capped at $1200$ tokens of cue-to-response heuristics) together
with the best-skill cache $m_{r^*}$ used for rollback. The
\emph{curve} is the score sequence $[S_0, S_1, \ldots, S_r]$
accumulated so far.

\noindent\textbf{Best-skill rollback.}
\label{sec:method:rollback}
Reflection is not monotone. When a round's score drops sharply
below the best seen so far ($S_{r+1} < \alpha \cdot S_{r^*}$,
$\alpha = 0.5$), the harness resets the next round's starting
prompt to $m_{r^*}$ and resumes reflection from there, guarding
against catastrophic skill drift.

\noindent\textbf{The curve.}
Iterating for $R$ rounds yields $[S_0, S_1, \ldots, S_R]$, the
\emph{Improvement Dynamics Curve} of the (agent, game) pair. Two
models with the same final $S_R$ can produce very different
curves (early vs.\ late convergence, monotone vs.\ oscillating);
the curve, rather than any single round, is the measurement
object in \S\ref{sec:experiments:idc}.

\section{Experiments}
\label{sec:experiments}

We report results in three blocks. \S\ref{sec:experiments:settings}
fixes the evaluation protocol, the set of agents, and the scoring
conventions used throughout. \S\ref{sec:experiments:main} reports
the main \emph{cold-start} leaderboard, in which every agent
plays every game once with no prior trajectories and no provided
skill, spanning the seven Solo, three PvP, and two Coop games.
\S\ref{sec:experiments:idc} relaxes the cold-start constraint via
the \emph{Improvement Dynamics Curve} (IDC), an agentic
self-reflection protocol in which the same model alternates
between playing the game and rewriting its own \emph{skill} (a
natural-language summary of game-specific strategy) for $R$
rounds. We use IDC to characterize (i) how each model improves on
the original task, and (ii) whether the learned skill transfers
to three held-out task variants per game.

\subsection{Experimental Settings}
\label{sec:experiments:settings}

\noindent\textbf{Agents.} 
We evaluate twelve agents grouped into three
classes so their strengths and limitations are not averaged away.
\textbf{(a) Commercial VLMs (API-only):} three Claude models
(Opus~4.7~\citep{anthropic2026opus47},
Opus~4.6~\citep{anthropic2026opus46},
Sonnet~4.6~\citep{anthropic2026sonnet46}), two OpenAI models
(GPT-5.5~\citep{openai2026gpt55} and
GPT-5.4~\citep{openai2026gpt54}), two Gemini models (3.1~Pro~Preview
and 3.1~Flash-Lite~Preview~\citep{google2026gemini31pro}), and
Kimi~K2.5~\citep{kimi_k25_2026}.
\textbf{(b) Open-weight VLMs:} two Qwen3.5 mixture-of-experts
checkpoints, 397B-A17B and 122B-A10B~\citep{qwen2026qwen35}, served
locally behind an OpenAI-compatible endpoint.
\textbf{(c) Specialized game policies:}
\textsc{NitroGen}~\citep{magne2026nitrogen}, which consumes a single
frame and emits a chunk of 18 gamepad actions (21-dim each), and
\textsc{Open-P2P}~\citep{yue2026scaling}, which consumes a 200-frame
keyboard-mouse history and emits one action per frame. Both run with
the native real-time protocols from their original papers.
All agents are evaluated on the same OmniGameArena real-time
environment with matched configuration. 
VLMs use OmniGameArena's chunked keyboard-mouse adapter, while the specialized
policies are routed through their native interfaces unchanged, so
each system is evaluated at the operating point it was designed for.

\noindent\textbf{Evaluation protocol.} 
Each (agent, game) cell is
evaluated over $N{=}5$ episodes; PvP cells additionally cover every
pairwise matchup in the agent pool. OmniGameArena natively runs in
real time, but commercial VLM API calls suffer from network jitter
that is orthogonal to model capability. We therefore evaluate under
two clock modes that both pause the environment during inference:
Paused Decision Quality (PDQ) freezes the environment for
the full inference call and treats decision time as free, isolating
pure decision quality; Latency-Controlled Real-Time (LCRT)
additionally idles for the server-reported inference time before
each action is applied, charging agents for on-device latency while
excluding network round-trip noise. Main-table results use PDQ and
LCRT is reported in Appendix \ref{sec:lcrt}.

\begin{figure}[t]
    \centering
    \includegraphics[width=\linewidth]{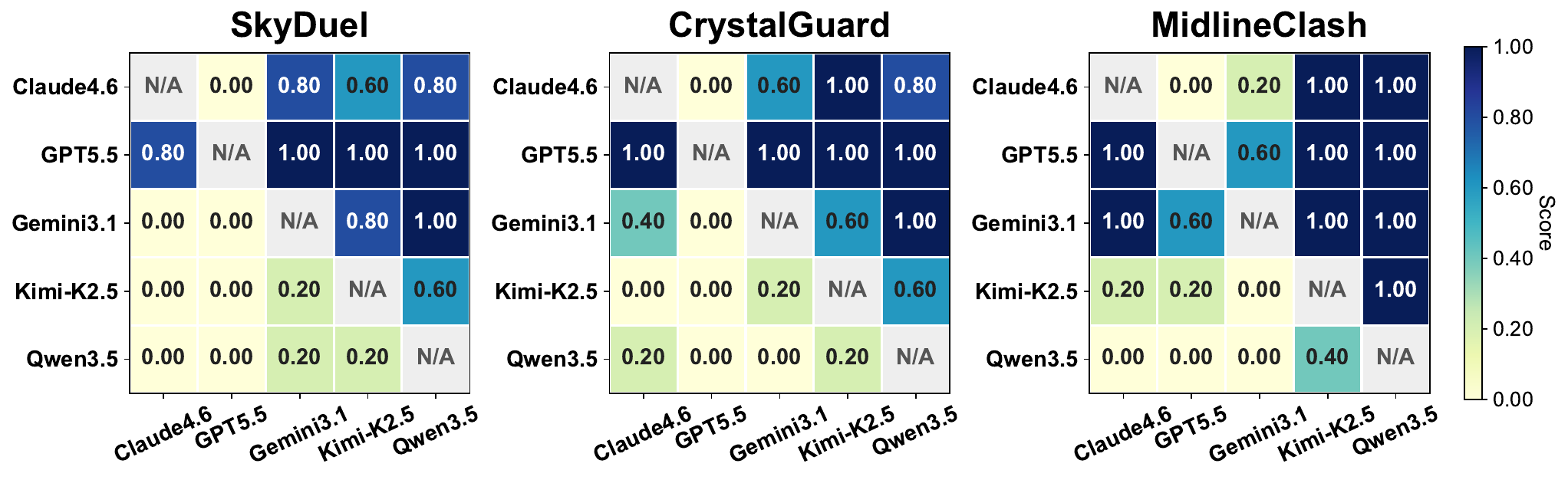}
    \caption{PvP win rates of Player~1 (row) against Player~2 (column) per game over all pairings.}
    \label{fig:omni_game_arena_heatmaps}
\end{figure}

\subsection{Cold-start Leaderboard}
\label{sec:experiments:main}
\noindent\textbf{Solo games.} 
Table~\ref{tab:main-solo} reports the seven Solo games. 
Three patterns stand out. \textbf{(1) No single model
dominates.} Leadership rotates across games: GPT-5.5 leads four of
seven games (ObstacleRun2D, LastStand, SceneEscape, SoloCraft),
Claude Opus~4.6 wins CueChase by a wide margin ($0.840$ vs.\ next
$0.580$), and Gemini~3.1 Pro leads MonsterShoot ($0.710$ vs.\ next
$0.464$). \textbf{(2) Newer is not always better.} Claude Opus~4.6
outperforms Opus~4.7 on five of seven Solo games, and GPT-5.4
exceeds GPT-5.5 on SceneEscape, indicating that capability ranking
is task-specific rather than monotone in release order.
\textbf{(3) Open-weight VLMs and specialized policies fail to
transfer.} Both Qwen3.5 MoE checkpoints score below $0.15$ on every
game and exactly $0.00$ on several, while
\textsc{NitroGen}~(gamepad) and \textsc{Open-P2P}~(keyboard-mouse)
collapse to near zero on all but a handful of games. This confirms
that OmniGameArena's task diversity lies well outside the training
distribution of policies optimized for narrow single-game.

\noindent\textbf{PvP games.} 
Figure~\ref{fig:omni_game_arena_heatmaps} reports Player~1 win
rates over all pairings (diagonal omitted). SkyDuel and CrystalGuard
show a clean dominance hierarchy that tracks the Solo leaderboard:
GPT-5.5 and Gemini~3.1 Pro win against nearly every opponent,
while both Qwen3.5 variants lose nearly every matchup.
MidlineClash is \textit{non-transitive}: Kimi~K2.5 beats
Claude~Opus~4.6 in all five Player~1 matches ($1.00$) even
though Claude is the stronger Solo agent, and Claude wins
decisively only against Qwen3.5. This indicates that MidlineClash
rewards game-specific tactics that do not align with the Solo
capability ranking.

\noindent\textbf{Coop games.} 
Table~\ref{tab:main-coop} reports team scores when two copies of
the same model cooperate. The leaderboard mirrors Solo: GPT-5.5
leads both games, Gemini~3.1 Pro is a close second, and Claude
Opus~4.6 follows. Two findings extend beyond Solo. First, the gap
between commercial and open-weight VLMs widens: both Qwen3.5
checkpoints score exactly $0.000$ on both games, failing to
complete a single shared order or hand-off. 
Second, even the strongest model reaches only $0.368$ on SharedFloor and $0.184$ on
HandoffRun, leaving substantial headroom and indicating that
LLM-LLM coordination is an unsolved capability gap rather than a
saturated benchmark dimension.

\begin{table}[t]
\centering
\setlength{\tabcolsep}{6pt}
\resizebox{1.0\linewidth}{!}{%
\begin{tabular}{lcc}
\toprule
Agent & SharedFloor & HandoffRun \\
\midrule
\modelicon{claude}Claude Opus 4.7         \citep{anthropic2026opus47}        
& {$0.136_{\pm 0.033}$} & {$0.040_{\pm 0.040}$} \\
\modelicon{claude}Claude Opus 4.6              \citep{anthropic2026opus46}   
& \cellcolor{third}{$0.152_{\pm 0.030}$} & \cellcolor{third}{$0.064_{\pm 0.036}$} \\
\modelicon{claude}Claude Sonnet 4.6            \citep{anthropic2026sonnet46}   
& {$0.148_{\pm 0.064}$} & {$0.008_{\pm 0.018}$} \\
\modelicon{openai}GPT-5.5      \citep{openai2026gpt55}                   
& \cellcolor{best}{$0.368_{\pm 0.036}$} & \cellcolor{best}{$0.184_{\pm 0.022}$} \\
\modelicon{openai}GPT-5.4         \citep{openai2026gpt54}             
& {$0.068_{\pm 0.023}$} & {$0.060_{\pm 0.024}$} \\
\modelicon{gemini}Gemini 3.1 Pro Preview       \citep{google2026gemini31pro}
& \cellcolor{second}{$0.336_{\pm 0.043}$} & \cellcolor{second}{$0.136_{\pm 0.043}$} \\
\modelicon{gemini}Gemini 3.1 Flash-Lite Preview \citep{google2026gemini31flashlite}
& {$0.052_{\pm 0.036}$} & {$0.020_{\pm 0.023}$} \\
\modelicon{kimi}Kimi K2.5                    \citep{kimi_k25_2026}  
& {$0.072_{\pm 0.023}$} & {$0.008_{\pm 0.018}$} \\
\midrule
\modelicon{qwen}Qwen3.5-397B-A17B              \citep{qwen2026qwen35}
& {$0.000_{\pm 0.000}$} & {$0.000_{\pm 0.000}$} \\
\modelicon{qwen}Qwen3.5-122B-A10B               \citep{qwen2026qwen35}
& {$0.008_{\pm 0.018}$} & {$0.000_{\pm 0.000}$} \\
\bottomrule
\end{tabular}%
}
\caption{Coop team scores on two cooperative games, 
where two copies of the same model must coordinate to complete a shared task. 
}
\label{tab:main-coop}
\end{table}

\begin{table*}[t]
\centering
\setlength{\tabcolsep}{4pt}
\resizebox{1.0\linewidth}{!}{%
\begin{tabular}{lcccccccc}
\toprule
& \multicolumn{4}{c}{LastStand} & \multicolumn{4}{c}{SharedFloor} \\
\cmidrule(lr){2-5} \cmidrule(lr){6-9}
Agent & origin & var1 & var2 & var3 & origin & var1 & var2 & var3 \\
\midrule
\modelicon{claude}Claude Opus 4.6
& \cellcolor{gainstrong}{$+0.641\ (+437\%)$}
& \cellcolor{gainstrong}{$+0.279\ (+175\%)$}
& \cellcolor{lossstrong}{$-0.168\ (-72\%)$}
& \cellcolor{gainmild}{$+0.011\ (+6\%)$}
& \cellcolor{gainmid}{$+0.060\ (+39\%)$}
& \cellcolor{gainstrong}{$+0.036\ (+56\%)$}
& \cellcolor{gainmid}{$+0.036\ (+38\%)$}
& \cellcolor{gainmid}{$+0.028\ (+30\%)$} \\
\modelicon{claude}Claude Opus 4.7
& \cellcolor{gainstrong}{$+0.620\ (+201\%)$}
& \cellcolor{lossmid}{$-0.097\ (-45\%)$}
& \cellcolor{lossstrong}{$-0.266\ (-76\%)$}
& \cellcolor{lossmid}{$-0.044\ (-21\%)$}
& \cellcolor{gainstrong}{$+0.068\ (+50\%)$}
& \cellcolor{gainmid}{$+0.040\ (+43\%)$}
& \cellcolor{gainmid}{$+0.020\ (+17\%)$}
& \cellcolor{gainmid}{$+0.024\ (+22\%)$} \\
\modelicon{openai}GPT-5.5
& \cellcolor{gainstrong}{$+0.540\ (+130\%)$}
& \cellcolor{gainstrong}{$+0.292\ (+57\%)$}
& \cellcolor{gainstrong}{$+0.422\ (+79\%)$}
& \cellcolor{gainmild}{$+0.012\ (+6\%)$}
& \cellcolor{gainmild}{$+0.020\ (+5\%)$}
& \cellcolor{gainmid}{$+0.034\ (+13\%)$}
& \cellcolor{gainmild}{$+0.016\ (+6\%)$}
& \cellcolor{gainmild}{$+0.020\ (+10\%)$} \\
\modelicon{gemini}Gemini 3.1 Pro Preview
& \cellcolor{gainstrong}{$+0.701\ (+305\%)$}
& \cellcolor{gainstrong}{$+0.266\ (+166\%)$}
& \cellcolor{lossmid}{$-0.062\ (-36\%)$}
& \cellcolor{gainmid}{$+0.017\ (+12\%)$}
& \cellcolor{gainmid}{$+0.076\ (+23\%)$}
& \cellcolor{gainmid}{$+0.040\ (+17\%)$}
& \cellcolor{gainmild}{$+0.016\ (+6\%)$}
& \cellcolor{gainmild}{$+0.020\ (+9\%)$} \\
\bottomrule
\end{tabular}%
}
\caption{Transfer of IDC best skill across the origin split and three unseen variants.
Each cell shows the gain $V_{\text{best}}-V_0$ followed by the $\%$-change relative to $V_0$.
$V_0$ is the no-skill baseline on that split; $V_{\text{best}}$ applies the skill that gave the highest mean score during the 10-round run.
}
\label{tab:idc-transfer}
\end{table*}

\subsection{Improvement Dynamics Curve}
\label{sec:experiments:idc}

\noindent\textbf{Setup.} 
We run IDC on \textsc{LastStand} (Solo) and \textsc{SharedFloor}
(Coop) for complementary skill coverage: \textsc{LastStand}
stresses reactive control as tiles progressively fall away, while
\textsc{SharedFloor} combines explicit task rules with
multi-agent coordination. We exclude PvP to keep IDC gains
attributable to the reflector rather than to opponent behavior.
Four top-performing agents from the cold-start leaderboard
participate: Claude Opus~4.6, Claude Opus~4.7, GPT-5.5, and
Gemini~3.1 Pro. Each agent completes $R{=}10$ rounds of $K{=}5$
episodes under PDQ; Round~$0$ reuses the cold-start baseline from
\S\ref{sec:experiments:main}. The best skill from each (model,
game) run is then reapplied to three held-out task variants per
game.

\begin{figure}[t]
    \centering
    \includegraphics[width=\linewidth]{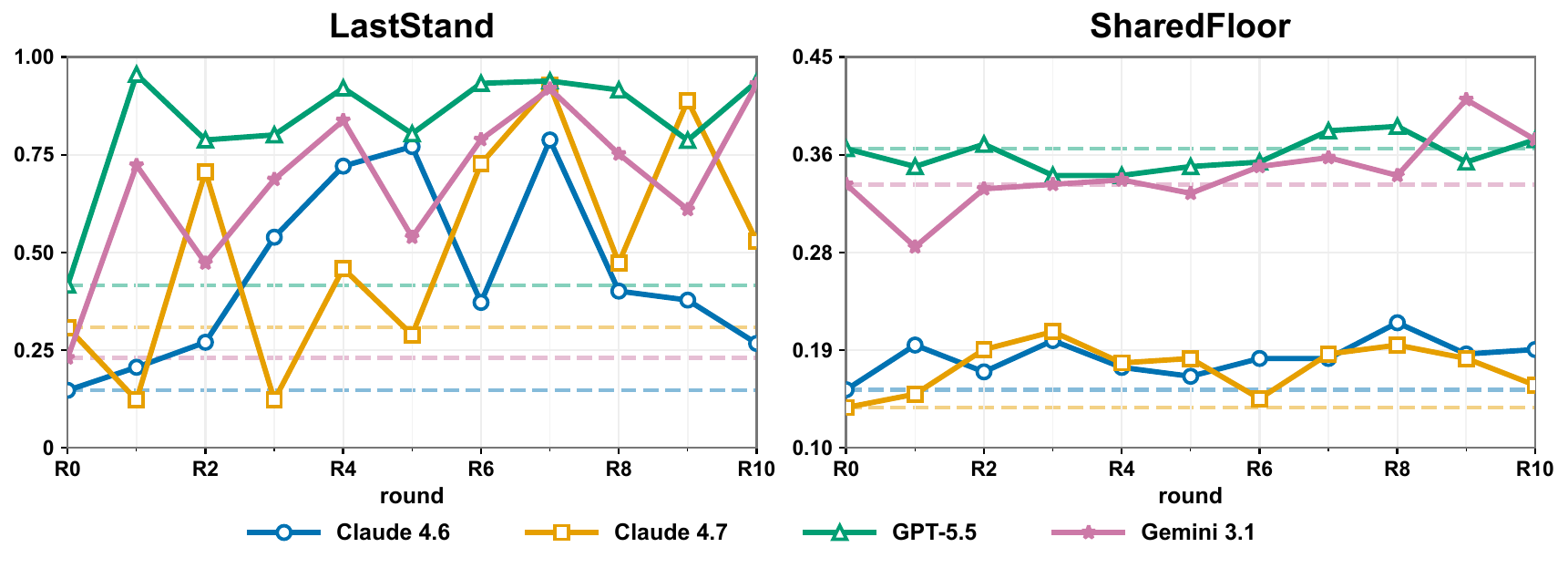}
    \caption{IDC curves: per-round mean episode score across 10 reflection rounds for four agents on LastStand (left) and SharedFloor (right). Horizontal lines mark each agent's R0 (no-skill PDQ baseline) for reference; deviations above the line indicate skill-driven improvement. Each round aggregates 5 episodes per agent.}
    \label{fig:idc-curve}
\end{figure}

\subsubsection{Improvement on the original task}
\label{sec:exp:idc-curve}

Figure~\ref{fig:idc-curve} reports per-round mean score. Through
multi-round agentic reflection, every model on both games
discovers skills that improve over its Round~0 baseline. On
LastStand, best-round gains range from $+0.54$ to $+0.70$ in
survival fraction ($+130\%$ to $+437\%$ relative). On SharedFloor,
best-round teams complete $1$ to $4$ additional orders per
episode, a substantial improvement in coordination throughput.
However, peak performance is typically reached mid-curve rather
than at Round~10: on LastStand the two Opus models lose $0.40$ to
$0.52$ between best and final round, while GPT-5.5 and Gemini
retain almost all gains. This best-vs-final gap justifies the
best-skill rollback mechanism described in
\S\ref{sec:method:rollback}.

\subsubsection{Transfer to held-out variants}
\label{sec:exp:idc-transfer}

Table~\ref{tab:idc-transfer} reports the gain of each best skill
on the origin split and three held-out variants per game.
\textbf{SharedFloor transfers universally ($16/16$ positive
across variants).} Gains range from $+6\%$ to $+56\%$,
corresponding to roughly $1$ to $2$ additional completed orders
per episode under the best skill. The skills encode coordination
heuristics rather than spatial memory: agents are guided to
observe the teammate's position and currently held item before
committing to an order (to avoid both players picking up the
same item), to discard duplicates at the trash bin when this
happens, to keep distance from the teammate's workstation to
prevent crowding, and to maintain a visible division of labor
across stations. Because the variants only change workbench and
item placements while preserving coordination rules, these
behavioral heuristics transfer without modification.
\textbf{LastStand transfer depends on skill style, not origin
gain magnitude.} The origin drops tiles one at a time, so three
of four models converge to a ``find a safe tile and stay''
policy that exploits the single-tile structure. \textsc{var1}
(different seed, same mechanics) leaves the structure intact and
transfers positively for three of four models. 
\textsc{var2} (cluster drops of multiple connected tiles) removes
the safe pocket the skills relied on: both Opus models collapse
($-72\%$ and $-76\%$), while \emph{GPT-5.5 gains $+79\%$}. Skill
inspection (Appendix~\ref{app:idc-skills}) explains the
divergence: the Opus and Gemini skills converge on
\emph{movement-minimizing} policies (e.g., ``stand still unless
your tile is red'', ``never chain forward steps''), which are
optimal under single-tile drops but maladaptive when a cluster
wipes out the safe pocket. GPT-5.5's skill instead encodes a
``move briefly, then reassess'' loop that adapts to either
mechanic.
\textsc{var3} (tiles that
track the player) eliminates static safety entirely; transfer
drops to small or negative gains for all four models. GPT-5.5
is the only model with positive transfer on all three variants
yet has the smallest origin gain ($+130\%$); Opus~4.7 gains
$+201\%$ on origin but transfers negatively on every variant.
This dissociation between origin gain and transferability is the
central finding of the variant experiment.

\section{Conclusion}
\label{sec:conclusion}

We introduced OmniGameArena, a benchmark of twelve newly built
UE5 real-time games spanning Solo, PvP, and Coop, and the
Improvement Dynamics Curve (IDC), an agentic-reflection harness
that produces multi-round self-improvement trajectories. Beyond
single-round leaderboard scores, the IDC exposes two additional
observables for each (agent, game) pair: how the score evolves
across reflection rounds, and how the learned skill behaves on
held-out task variants.

\section*{Limitations}
\label{sec:limitations}

\noindent\textbf{IDC scope.}
Due to compute constraints, our IDC experiments cover only two
environments (\textsc{LastStand} and \textsc{SharedFloor}), each
with three held-out variants, and four agents from the cold-start
leaderboard. Scaling IDC to additional games, variants, and
models is an extension.

\noindent\textbf{Single-skill format.}
Our reflector maintains a single bounded skill prompt that is
replaced each round, rather than a growing library of skills in
the style of Voyager. Library-based extensions are orthogonal
to the round-by-round refinement.

\noindent\textbf{Shared model for player and reflector.}
Each agent uses the same underlying model as both player and
reflector. Whether asymmetric setups (e.g., a smaller player
paired with a stronger reflector) yield different improvement is untested.

\bibliography{custom}

\begin{thebibliography}{39}
\providecommand{\natexlab}[1]{#1}

\bibitem[{{Anthropic}(2026{\natexlab{a}})}]{anthropic2026opus46}
{Anthropic}. 2026{\natexlab{a}}.
\newblock Introducing {C}laude {O}pus 4.6.
\newblock \url{https://www.anthropic.com/news/claude-opus-4-6}.

\bibitem[{{Anthropic}(2026{\natexlab{b}})}]{anthropic2026opus47}
{Anthropic}. 2026{\natexlab{b}}.
\newblock Introducing {C}laude {O}pus 4.7.
\newblock \url{https://www.anthropic.com/news/claude-opus-4-7}.

\bibitem[{{Anthropic}(2026{\natexlab{c}})}]{anthropic2026sonnet46}
{Anthropic}. 2026{\natexlab{c}}.
\newblock Introducing {C}laude {S}onnet 4.6.
\newblock \url{https://www.anthropic.com/news/claude-sonnet-4-6}.

\bibitem[{Bai et~al.(2026)Bai, Taymanov, Zhang, Kumar, and Whitehead}]{bai2026webgym}
Hao Bai, Alexey Taymanov, Tong Zhang, Aviral Kumar, and Spencer Whitehead. 2026.
\newblock Webgym: Scaling training environments for visual web agents with realistic tasks.
\newblock \emph{arXiv preprint arXiv:2601.02439}.

\bibitem[{Fan et~al.(2022)Fan, Wang, Jiang, Mandlekar, Yang, Zhu, Tang, Huang, Zhu, and Anandkumar}]{fan2022minedojo}
Linxi Fan, Guanzhi Wang, Yunfan Jiang, Ajay Mandlekar, Yuncong Yang, Haoyi Zhu, Andrew Tang, De-An Huang, Yuke Zhu, and Anima Anandkumar. 2022.
\newblock Minedojo: Building open-ended embodied agents with internet-scale knowledge.
\newblock volume~35, pages 18343--18362.

\bibitem[{Feng et~al.(2023)Feng, Luo, Wang, Tang, Yang, Shao, Mguni, Du, and Wang}]{feng2023chessgpt}
Xidong Feng, Yicheng Luo, Ziyan Wang, Hongrui Tang, Mengyue Yang, Kun Shao, David Mguni, Yali Du, and Jun Wang. 2023.
\newblock Chessgpt: Bridging policy learning and language modeling.
\newblock \emph{Advances in Neural Information Processing Systems}, 36:7216--7262.

\bibitem[{{Google}(2026{\natexlab{a}})}]{google2026gemini31flashlite}
{Google}. 2026{\natexlab{a}}.
\newblock Introducing {G}emini 3.1 {F}lash-{L}ite.
\newblock \url{https://blog.google/innovation-and-ai/models-and-research/gemini-models/gemini-3-1-flash-lite/}.

\bibitem[{{Google}(2026{\natexlab{b}})}]{google2026gemini31pro}
{Google}. 2026{\natexlab{b}}.
\newblock Introducing {G}emini 3.1 {P}ro.
\newblock \url{https://blog.google/innovation-and-ai/models-and-research/gemini-models/gemini-3-1-pro/}.

\bibitem[{Hausknecht et~al.(2020)Hausknecht, Ammanabrolu, C{\^o}t{\'e}, and Yuan}]{hausknecht2020interactive}
Matthew Hausknecht, Prithviraj Ammanabrolu, Marc-Alexandre C{\^o}t{\'e}, and Xingdi Yuan. 2020.
\newblock Interactive fiction games: A colossal adventure.
\newblock In \emph{Proceedings of the AAAI Conference on Artificial Intelligence}, volume~34, pages 7903--7910.

\bibitem[{Hu et~al.(2025)Hu, Huo, Zhang, Yu, Xing, Stoica, Rosing, Jin, and Zhang}]{hu2025lmgamebench}
Lanxiang Hu, Mingjia Huo, Yuxuan Zhang, Haoyang Yu, Eric~P Xing, Ion Stoica, Tajana Rosing, Haojian Jin, and Hao Zhang. 2025.
\newblock lmgame-bench: How good are llms at playing games?
\newblock \emph{arXiv preprint arXiv:2505.15146}.

\bibitem[{Hu et~al.(2024)Hu, Li, Xie, Jiang, Stoica, Jin, and Zhang}]{hu2024gamearena}
Lanxiang Hu, Qiyu Li, Anze Xie, Nan Jiang, Ion Stoica, Haojian Jin, and Hao Zhang. 2024.
\newblock Gamearena: Evaluating llm reasoning through live computer games.
\newblock \emph{arXiv preprint arXiv:2412.06394}.

\bibitem[{Huang et~al.(2024)Huang, Li, Lam, Liang, Wang, Yuan, Jiao, Wang, Tu, and Lyu}]{huang2024far}
Jen-tse Huang, Eric~John Li, Man~Ho Lam, Tian Liang, Wenxuan Wang, Youliang Yuan, Wenxiang Jiao, Xing Wang, Zhaopeng Tu, and Michael~R Lyu. 2024.
\newblock How far are we on the decision-making of llms? evaluating llms' gaming ability in multi-agent environments.
\newblock \emph{arXiv preprint arXiv:2403.11807}.

\bibitem[{K{\"u}ttler et~al.(2020)K{\"u}ttler, Nardelli, Miller, Raileanu, Selvatici, Grefenstette, and Rockt{\"a}schel}]{kuttler2020nethack}
Heinrich K{\"u}ttler, Nantas Nardelli, Alexander Miller, Roberta Raileanu, Marco Selvatici, Edward Grefenstette, and Tim Rockt{\"a}schel. 2020.
\newblock The nethack learning environment.
\newblock \emph{Advances in Neural Information Processing Systems}, 33:7671--7684.

\bibitem[{Li et~al.(2025)Li, Wang, He, Ma, and Liang}]{li2025jarvis}
Muyao Li, Zihao Wang, Kaichen He, Xiaojian Ma, and Yitao Liang. 2025.
\newblock Jarvis-vla: Post-training large-scale vision language models to play visual games with keyboards and mouse.
\newblock In \emph{Findings of the Association for Computational Linguistics: ACL 2025}, pages 17878--17899.

\bibitem[{Lin et~al.(2025)Lin, Huang, Li, Jiang, Wu, Zhong, Qian, Wang, and Qi}]{lin2025embrace}
Mingxian Lin, Wei Huang, Yitang Li, Chengjie Jiang, Kui Wu, Fangwei Zhong, Shengju Qian, Xin Wang, and Xiaojuan Qi. 2025.
\newblock Embrace-3k: Embodied reasoning and action in complex environments.
\newblock \emph{arXiv preprint arXiv:2507.10548}.

\bibitem[{Luo et~al.(2025)Luo, Zhao, Lin, Zhu, Tang, Liu, Chen, Qian, Wang, and You}]{luo2025v}
Yang Luo, Xuanlei Zhao, Baijiong Lin, Lingting Zhu, Liyao Tang, Yuqi Liu, Ying-Cong Chen, Shengju Qian, Xin Wang, and Yang You. 2025.
\newblock V-reasonbench: Toward unified reasoning benchmark suite for video generation models.
\newblock \emph{arXiv preprint arXiv:2511.16668}.

\bibitem[{Madaan et~al.(2023)Madaan, Tandon, Gupta, Hallinan, Gao, Wiegreffe, Alon, Dziri, Prabhumoye, Yang et~al.}]{madaan2023selfrefine}
Aman Madaan, Niket Tandon, Prakhar Gupta, Skyler Hallinan, Luyu Gao, Sarah Wiegreffe, Uri Alon, Nouha Dziri, Shrimai Prabhumoye, Yiming Yang, and 1 others. 2023.
\newblock Self-refine: Iterative refinement with self-feedback.
\newblock \emph{Advances in neural information processing systems}, 36:46534--46594.

\bibitem[{Magne et~al.(2026)Magne, Awadalla, Wang, Xu, Belofsky, Hu, Kim, Schmidt, Gkioxari, Kautz et~al.}]{magne2026nitrogen}
Lo{\"\i}c Magne, Anas Awadalla, Guanzhi Wang, Yinzhen Xu, Joshua Belofsky, Fengyuan Hu, Joohwan Kim, Ludwig Schmidt, Georgia Gkioxari, Jan Kautz, and 1 others. 2026.
\newblock Nitrogen: An open foundation model for generalist gaming agents.
\newblock \emph{arXiv preprint arXiv:2601.02427}.

\bibitem[{{Moonshot AI}(2026)}]{kimi_k25_2026}
{Moonshot AI}. 2026.
\newblock Kimi k2.5: Visual agentic intelligence.
\newblock \url{https://www.kimi.com/blog/kimi-k2-5}.

\bibitem[{{OpenAI}(2026{\natexlab{a}})}]{openai2026gpt54}
{OpenAI}. 2026{\natexlab{a}}.
\newblock Introducing {GPT}-5.4.
\newblock \url{https://openai.com/index/introducing-gpt-5-4/}.

\bibitem[{{OpenAI}(2026{\natexlab{b}})}]{openai2026gpt55}
{OpenAI}. 2026{\natexlab{b}}.
\newblock Introducing {GPT}-5.5.
\newblock \url{https://openai.com/index/introducing-gpt-5-5/}.

\bibitem[{Paglieri et~al.(2024)Paglieri, Cupia{\l}, Coward, Piterbarg, Wolczyk, Khan, Pignatelli, Kuci{\'n}ski, Pinto, Fergus et~al.}]{paglieri2024balrog}
Davide Paglieri, Bart{\l}omiej Cupia{\l}, Samuel Coward, Ulyana Piterbarg, Maciej Wolczyk, Akbir Khan, Eduardo Pignatelli, {\L}ukasz Kuci{\'n}ski, Lerrel Pinto, Rob Fergus, and 1 others. 2024.
\newblock Balrog: Benchmarking agentic llm and vlm reasoning on games.
\newblock \emph{arXiv preprint arXiv:2411.13543}.

\bibitem[{Park et~al.(2025)Park, Kim, Choi, Kim, Lee, Lee, Park, Lee, Hwang, Ahn et~al.}]{park2025orak}
Dongmin Park, Minkyu Kim, Beongjun Choi, Junhyuck Kim, Keon Lee, Jonghyun Lee, Inkyu Park, Byeong-Uk Lee, Jaeyoung Hwang, Jaewoo Ahn, and 1 others. 2025.
\newblock Orak: A foundational benchmark for training and evaluating llm agents on diverse video games.
\newblock \emph{arXiv preprint arXiv:2506.03610}.

\bibitem[{{Qwen Team}(2026)}]{qwen2026qwen35}
{Qwen Team}. 2026.
\newblock {Qwen3.5}: A {N}ative {M}ultimodal {F}oundation {M}odel for {E}fficiency.
\newblock \url{https://qwen.ai/blog?id=qwen3.5}.

\bibitem[{Shinn et~al.(2023)Shinn, Cassano, Gopinath, Narasimhan, and Yao}]{shinn2023reflexion}
Noah Shinn, Federico Cassano, Ashwin Gopinath, Karthik Narasimhan, and Shunyu Yao. 2023.
\newblock Reflexion: Language agents with verbal reinforcement learning.
\newblock \emph{Advances in neural information processing systems}, 36:8634--8652.

\bibitem[{Tan et~al.(2024)Tan, Ding, Zhang, Li, Zhou, Yue, Xia, Jiang, Zheng, Xu et~al.}]{tan2024cradle}
Weihao Tan, Ziluo Ding, Wentao Zhang, Boyu Li, Bohan Zhou, Junpeng Yue, Haochong Xia, Jiechuan Jiang, Longtao Zheng, Xinrun Xu, and 1 others. 2024.
\newblock Towards general computer control: A multimodal agent for red dead redemption ii as a case study.
\newblock \emph{arXiv preprint arXiv:2403.03186}, 1(2).

\bibitem[{Tan et~al.(2025)Tan, Li, Fang, Yao, Yan, Luo, Ao, Li, Ren, Yi et~al.}]{tan2025lumine}
Weihao Tan, Xiangyang Li, Yunhao Fang, Heyuan Yao, Shi Yan, Hao Luo, Tenglong Ao, Huihui Li, Hongbin Ren, Bairen Yi, and 1 others. 2025.
\newblock Lumine: An open recipe for building generalist agents in 3d open worlds.
\newblock \emph{arXiv preprint arXiv:2511.08892}.

\bibitem[{Tsai et~al.(2023)Tsai, Zhou, Liu, Li, Yu, and Mei}]{tsai2023can}
Chen~Feng Tsai, Xiaochen Zhou, Sierra~S Liu, Jing Li, Mo~Yu, and Hongyuan Mei. 2023.
\newblock Can large language models play text games well? current state-of-the-art and open questions.
\newblock \emph{arXiv preprint arXiv:2304.02868}.

\bibitem[{Wang et~al.(2023)Wang, Xie, Jiang, Mandlekar, Xiao, Zhu, Fan, and Anandkumar}]{wang2023voyager}
Guanzhi Wang, Yuqi Xie, Yunfan Jiang, Ajay Mandlekar, Chaowei Xiao, Yuke Zhu, Linxi Fan, and Anima Anandkumar. 2023.
\newblock Voyager: An open-ended embodied agent with large language models, 2023.
\newblock \emph{URL https://arxiv. org/abs/2305.16291}, 2(11).

\bibitem[{Wang et~al.(2025{\natexlab{a}})Wang, Zhuang, and Wu}]{wang2025large}
Xinyu Wang, Bohan Zhuang, and Qi~Wu. 2025{\natexlab{a}}.
\newblock Are large vision language models good game players?
\newblock \emph{arXiv preprint arXiv:2503.02358}.

\bibitem[{Wang et~al.(2025{\natexlab{b}})Wang, Li, Ye, Fang, Wang, Liu, Liang, Lu, Wu, Feng et~al.}]{wang2025game}
Zihao Wang, Xujing Li, Yining Ye, Junjie Fang, Haoming Wang, Longxiang Liu, Shihao Liang, Junting Lu, Zhiyong Wu, Jiazhan Feng, and 1 others. 2025{\natexlab{b}}.
\newblock Game-tars: Pretrained foundation models for scalable generalist multimodal game agents.
\newblock \emph{arXiv preprint arXiv:2510.23691}.

\bibitem[{Weng(2026)}]{weng2026learning_beyond_gradients}
Jiayi Weng. 2026.
\newblock Learning beyond gradients.
\newblock \url{https://trinkle23897.github.io/learning-beyond-gradients/}.
\newblock Blog post.

\bibitem[{Wu et~al.(2023)Wu, Tang, Mitchell, and Li}]{wu2023smartplay}
Yue Wu, Xuan Tang, Tom~M Mitchell, and Yuanzhi Li. 2023.
\newblock Smartplay: A benchmark for llms as intelligent agents.
\newblock \emph{arXiv preprint arXiv:2310.01557}.

\bibitem[{Yue et~al.(2026)Yue, Salia, Hunt, Green, Shi, and Hunt}]{yue2026scaling}
Yuguang Yue, Irakli Salia, Samuel Hunt, Chris Green, Wenzhe Shi, and Jonathan~J Hunt. 2026.
\newblock Scaling behavior cloning improves causal reasoning: An open model for real-time video game playing.
\newblock \emph{arXiv preprint arXiv:2601.04575}.

\bibitem[{Zhang et~al.(2025)Zhang, Griffiths, Narasimhan, and Press}]{zhang2025videogamebench}
Alex~L Zhang, Thomas~L Griffiths, Karthik~R Narasimhan, and Ofir Press. 2025.
\newblock Videogamebench: Can vision-language models complete popular video games?
\newblock \emph{arXiv preprint arXiv:2505.18134}.

\bibitem[{Zhang et~al.(2026)Zhang, Liu, Zhao, Hou, Zhang, Xie, Liu, and Li}]{zhang2026gameverse}
Kuan Zhang, Dongchen Liu, Qiyue Zhao, Jinkun Hou, Xinran Zhang, Qinlei Xie, Miao Liu, and Yiming Li. 2026.
\newblock Gameverse: Can vision-language models learn from video-based reflection?
\newblock \emph{arXiv preprint arXiv:2603.06656}.

\bibitem[{Zhao et~al.(2024)Zhao, Huang, Xu, Lin, Liu, and Huang}]{zhao2024expel}
Andrew Zhao, Daniel Huang, Quentin Xu, Matthieu Lin, Yong-Jin Liu, and Gao Huang. 2024.
\newblock Expel: Llm agents are experiential learners.
\newblock In \emph{Proceedings of the AAAI Conference on Artificial Intelligence}, volume~38, pages 19632--19642.

\bibitem[{Zheng et~al.(2025)Zheng, Li, Yang, Yu, Wang, Yan, Yao, and Wang}]{zheng2025v}
Xiangxi Zheng, Linjie Li, Zhengyuan Yang, Ping Yu, Alex~Jinpeng Wang, Rui Yan, Yuan Yao, and Lijuan Wang. 2025.
\newblock V-mage: A game evaluation framework for assessing vision-centric capabilities in multimodal large language models.
\newblock \emph{arXiv preprint arXiv:2504.06148}.

\bibitem[{Zhu et~al.(2026)Zhu, Qian, Fan, Dong, Jin, Zhou, Dong, Wang, and Yu}]{zhu2026assetformer}
Lingting Zhu, Shengju Qian, Haidi Fan, Jiayu Dong, Zhenchao Jin, Siwei Zhou, Gen Dong, Xin Wang, and Lequan Yu. 2026.
\newblock Assetformer: Modular 3d assets generation with autoregressive transformer.
\newblock \emph{arXiv preprint arXiv:2602.12100}.

\end{thebibliography}

\clearpage
\appendix



\section{Latency-Controlled Real-Time Eval}
\label{sec:lcrt}

We further evaluate a subset of agents under a \textbf{Latency-Controlled
Real-Time} (LCRT) protocol. In PDQ, the simulator is paused while the model is
thinking, and the returned action is executed immediately from the observed
state. In LCRT, the simulator is likewise paused during the wall-clock model
call, but the model's measured decision latency is then injected back into the
game timeline before the action is executed. Thus an action predicted from
observation $o_t$ is applied after approximately $\Delta t$ seconds of simulated
game time, where $\Delta t$ is the model's decision latency.
LCRT requires a reliable estimate of \emph{pure model inference time}. We
therefore report LCRT only for the four models whose backends expose usable
model-side timing signals in our implementation: Claude Opus 4.6, Claude
Sonnet 4.6, GPT-5.5, and GPT-5.4. Other agents are excluded because their
available timings fold in client-side overheads such as request queuing,
network transfer, retries, or wrapper latency; injecting such end-to-end
wall-clock measurements would confound model latency with infrastructure
latency and make the comparison unfair.

\noindent\textbf{Solo Games.}
We focus the LCRT analysis on tasks where latency is expected to affect either
the evolving game state or the available time budget, and accordingly report
LastStand and MonsterShoot as dynamic real-time tasks and SoloCraft as a
time-budgeted interaction task. As shown in \Cref{tab:lcrt-solo}, the three
tasks reveal distinct sensitivity patterns. MonsterShoot is the most
latency-sensitive: all four models drop, most steeply GPT-5.5
($\Delta=-0.408$), consistent with its need for continuous aiming and target
tracking. LastStand behaves differently: its optimal policy is nearly
stationary, waiting on a safe tile and moving only when one's own tile is about
to fall, so latency is not necessarily harmful here, and taking fewer, later
actions can even avoid fatal missteps. This may explain why Claude Opus 4.6 and
GPT-5.4 even improve under LCRT and Claude Sonnet 4.6 stays essentially flat,
while GPT-5.5 is the only model that degrades. SoloCraft is not a
reactive-control task in the same sense; here latency mainly consumes the
episode time budget and reduces interaction throughput, an effect we quantify
below.

\begin{table}[t]
\centering
\small

\setlength{\tabcolsep}{3pt}
\resizebox{\linewidth}{!}{%
\begin{tabular}{lccc}
\toprule
Agent & LastStand & MonsterShoot & SoloCraft \\
\midrule

\modelicon{claude}~Claude Opus 4.6 \citep{anthropic2026opus46}
& \makecell{$0.248_{\pm 0.126}$\\{\scriptsize $\Delta$ +0.101}}
& \makecell{$0.180_{\pm 0.027}$\\{\scriptsize $\Delta$ -0.182}}
& \makecell{$0.080_{\pm 0.018}$\\{\scriptsize $\Delta$ -0.148}}
\\

\modelicon{claude}~Claude Sonnet 4.6 \citep{anthropic2026sonnet46}
& \makecell{$0.154_{\pm 0.058}$\\{\scriptsize $\Delta$ +0.010}}
& \makecell{$0.116_{\pm 0.033}$\\{\scriptsize $\Delta$ -0.050}}
& \makecell{$0.072_{\pm 0.010}$\\{\scriptsize $\Delta$ -0.052}}
\\

\modelicon{openai}~GPT-5.5  \citep{openai2026gpt55}
& \makecell{$0.324_{\pm 0.166}$\\{\scriptsize $\Delta$ -0.092}}
& \makecell{$0.056_{\pm 0.023}$\\{\scriptsize $\Delta$ -0.408}}
& \makecell{$0.052_{\pm 0.010}$\\{\scriptsize $\Delta$ -0.200}}
\\

\modelicon{openai}~GPT-5.4   \citep{openai2026gpt54}
& \makecell{$0.253_{\pm 0.094}$\\{\scriptsize $\Delta$ +0.105}}
& \makecell{$0.138_{\pm 0.107}$\\{\scriptsize $\Delta$ -0.188}}
& \makecell{$0.040_{\pm 0.036}$\\{\scriptsize $\Delta$ -0.044}}
\\

\bottomrule
\end{tabular}%
}
\caption{LCRT results on selected Solo games. Each cell reports mean with standard deviation as subscript over $N{=}5$ episodes, with $\Delta$ denoting the change relative to the corresponding PDQ result.}
\label{tab:lcrt-solo}
\end{table}

\begin{figure}[t]
    \centering
    \includegraphics[width=0.7\linewidth]{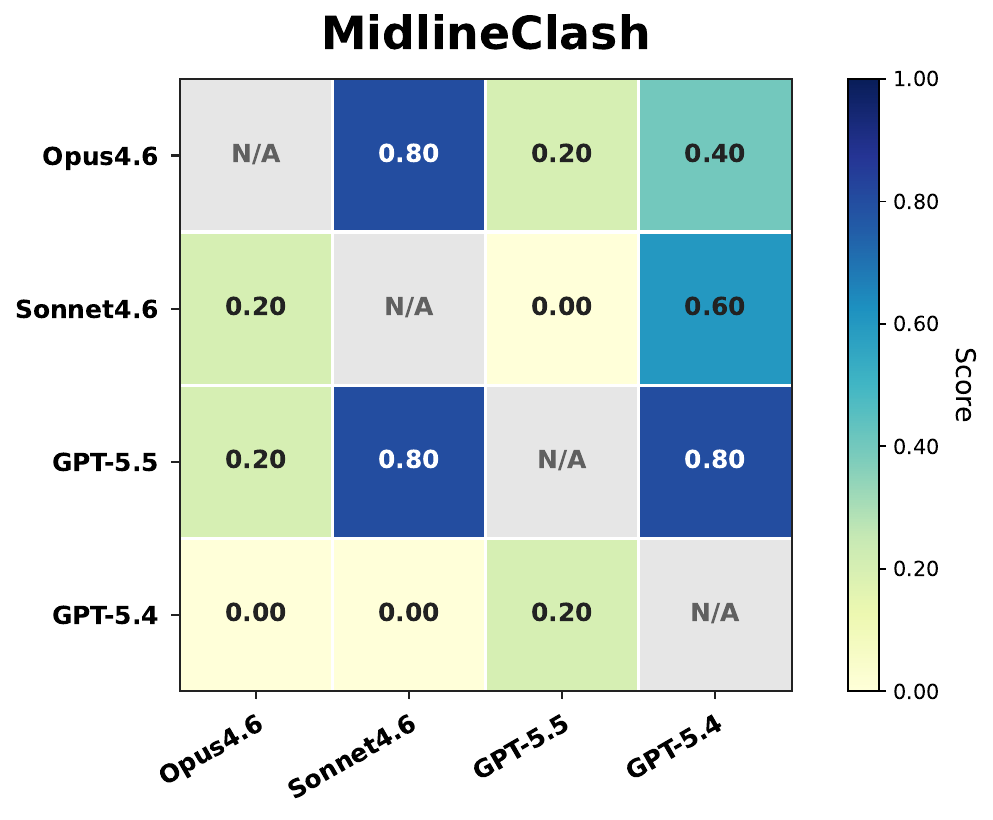}
    \caption{PvP win rates of Player~1 (row) against Player~2 (column) on MidlineClash under latency control setting.}
    \label{fig:midlineclash_heatmap}
\end{figure}

\noindent\textbf{PvP Games.}
\Cref{fig:midlineclash_heatmap} reports Player~1 win rates on MidlineClash under
LCRT for the four commercial VLMs. They differ from the PDQ heatmap
(\Cref{fig:omni_game_arena_heatmaps}), but the difference reflects the change of
clock mode rather than a change in model ability. Charging decision latency
against the game clock leaves far less game time per move, so each player
completes only about $18$ actions per game under LCRT against the $\sim$$42$-action
PDQ budget, and matches compress into low-scoring, frequently drawn games. The
effect is clearest in the GPT-5.5-vs-Opus~4.6 pairing, the one matchup both
protocols share: GPT-5.5 swept it $10$--$0$ under PDQ by wide margins (e.g.\
$8$--$0$, $10$--$4$, $11$--$1$), whereas under LCRT the same pairing yields $5$
GPT-5.5 wins, $3$ Opus wins, and $2$ draws, with narrow scorelines such as
$1$--$0$, $2$--$2$, and $0$--$0$ and the pairing's average per-player score
falling from $0.121$ to $0.016$. GPT-5.5 still wins the pairing and still posts
the highest average Player-1 win rate ($0.60$, versus $0.07$ for the weakest,
GPT-5.4), even though it carries by far the largest inference latency
($\sim$$16$~s of pure model time against $\sim$$4$--$7$~s for the others). Because
both sides pay the same delay, latency in symmetric play mainly compresses
margins and amplifies single-game randomness rather than re-ranking the agents,
a much milder effect than on the throughput tasks below.

\begin{table}[t]
\centering
\small
\setlength{\tabcolsep}{3pt}
\resizebox{\linewidth}{!}{%
\begin{tabular}{lc}
\toprule
Agent & SharedFloor \\
\midrule

\modelicon{claude}~Claude Opus 4.6 \citep{anthropic2026opus46}
& \makecell{$0.048_{\pm 0.016}$\\{\scriptsize $\Delta$ -0.104}}
\\

\modelicon{claude}~Claude Sonnet 4.6 \citep{anthropic2026sonnet46}
& \makecell{$0.028_{\pm 0.016}$\\{\scriptsize $\Delta$ -0.120}}
\\

\modelicon{openai}~GPT-5.5  \citep{openai2026gpt55}
& \makecell{$0.048_{\pm 0.020}$\\{\scriptsize $\Delta$ -0.320}}
\\

\modelicon{openai}~GPT-5.4   \citep{openai2026gpt54}
& \makecell{$0.024_{\pm 0.020}$\\{\scriptsize $\Delta$ -0.044}}
\\

\bottomrule
\end{tabular}%
}
\caption{LCRT results on the cooperative game SharedFloor. Each cell reports mean with standard deviation as subscript over $N{=}5$ episodes, with $\Delta$ denoting the change relative to the corresponding PDQ result.}
\label{tab:lcrt-coop}
\end{table}

\noindent\textbf{Cooperative Game.}
We further report the cooperative task SharedFloor, in which two instances of
the same model coordinate to complete shared orders before a fixed match
deadline (\Cref{tab:lcrt-coop}). Every model degrades under LCRT, and the loss
scales with the action budget: because LCRT charges each model's decision
latency against the match clock, the slowest agent completes the fewest
interactions. GPT-5.5, the slowest, takes the largest absolute drop, from
$0.368$ under PDQ to $0.048$ ($\Delta=-0.320$); but having started far ahead, it
still ties Opus~4.6 for the best LCRT score rather than collapsing. GPT-5.4, the
fastest, retains the most actions and changes the least ($\Delta=-0.044$). We
quantify this action-budget account next.

\noindent\textbf{Action budget versus per-action efficiency.}
To see why scores fall on the throughput tasks, we measure for each agent both
its actions per episode and its score per action under the two protocols on
SoloCraft (\Cref{tab:eff-solocraft}) and SharedFloor (\Cref{tab:eff-sharedfloor}).
The action budget shrinks monotonically with model speed: GPT-5.5, the slowest,
completes the fewest actions, only about $8$ of its $43$ PDQ actions per episode
on SoloCraft and $14$ versus $84$ on SharedFloor, while the faster agents retain
far more. The drop is overwhelmingly a budget effect: score per action is largely
preserved between PDQ and LCRT, most clearly for GPT-5.5 (essentially unchanged
on SoloCraft, only mildly lower on SharedFloor), so each agent's score falls
roughly in proportion to its smaller action count, e.g.\ GPT-5.5 on SoloCraft
drops $0.252\!\rightarrow\!0.052$ for $43\!\rightarrow\!8$ actions. This account
is specific to interaction-throughput tasks, where score accumulates with the
number of useful interactions before a fixed deadline; it does not extend to the
dynamic tasks, where latency acts through reaction timing rather than throughput
and is not even monotone, helping in LastStand, whose near-stationary optimal
policy rewards fewer and later actions, but hurting in MonsterShoot through
missed and mistimed shots. Together these regimes show that latency is a
first-class evaluation axis whose effect is task-dependent, taxing single-agent
throughput, compressing rather than re-ranking symmetric play, and even helping
near-stationary survival, so PDQ margins do not transfer directly to real-time
deployment.

\begin{table}[t]
\centering
\small
\setlength{\tabcolsep}{3pt}
\resizebox{\linewidth}{!}{%
\begin{tabular}{lcccc}
\toprule
& \multicolumn{2}{c}{Actions / episode} & \multicolumn{2}{c}{Score / action} \\
\cmidrule(lr){2-3}\cmidrule(lr){4-5}
Agent & PDQ & LCRT & PDQ & LCRT \\
\midrule
\modelicon{claude}~Claude Opus 4.6 \citep{anthropic2026opus46}   & 43 & 17 & 0.0053 & 0.0047 \\
\modelicon{claude}~Claude Sonnet 4.6 \citep{anthropic2026sonnet46} & 43 & 19 & 0.0029 & 0.0038 \\
\modelicon{openai}~GPT-5.5  \citep{openai2026gpt55}           & 43 & 8  & 0.0059 & 0.0065 \\
\modelicon{openai}~GPT-5.4   \citep{openai2026gpt54}           & 43 & 28 & 0.0020 & 0.0014 \\
\bottomrule
\end{tabular}%
}
\caption{\textbf{SoloCraft action budget.} Actions per episode and normalized score per action under PDQ vs.\ LCRT. Score per action is nearly unchanged, so each agent's LCRT drop (\Cref{tab:lcrt-solo}) is almost entirely its smaller action budget.}
\label{tab:eff-solocraft}
\end{table}

\begin{table}[t]
\centering
\small
\setlength{\tabcolsep}{3pt}
\resizebox{\linewidth}{!}{%
\begin{tabular}{lcccc}
\toprule
& \multicolumn{2}{c}{Actions / episode} & \multicolumn{2}{c}{Score / action} \\
\cmidrule(lr){2-3}\cmidrule(lr){4-5}
Agent & PDQ & LCRT & PDQ & LCRT \\
\midrule
\modelicon{claude}~Claude Opus 4.6 \citep{anthropic2026opus46}   & 84 & 33 & 0.0018 & 0.0015 \\
\modelicon{claude}~Claude Sonnet 4.6 \citep{anthropic2026sonnet46} & 84 & 36 & 0.0018 & 0.0008 \\
\modelicon{openai}~GPT-5.5  \citep{openai2026gpt55}           & 84 & 14 & 0.0044 & 0.0034 \\
\modelicon{openai}~GPT-5.4   \citep{openai2026gpt54}           & 84 & 55 & 0.0008 & 0.0004 \\
\bottomrule
\end{tabular}%
}
\caption{\textbf{SharedFloor action budget} (actions and team score summed over both players; score normalized). The LCRT drop tracks the shrinking action budget: GPT-5.5, the slowest, completes by far the fewest actions, yet keeps the highest score per action under both protocols, so its loss is a budget effect rather than a per-action collapse.}
\label{tab:eff-sharedfloor}
\end{table}


\begin{figure*}[t]
    \centering
    \includegraphics[width=\linewidth]{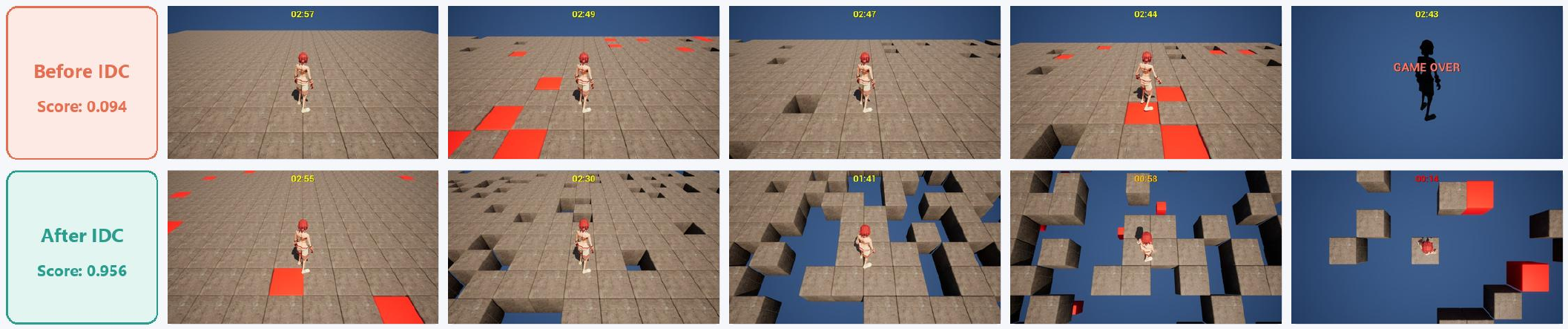}
    \caption{Qualitative comparison on \textit{Last Stand} using GPT-5.5.}
    \label{fig:last-stand-before-after}
\end{figure*}

\begin{figure*}[t]
    \centering
    \includegraphics[width=\linewidth]{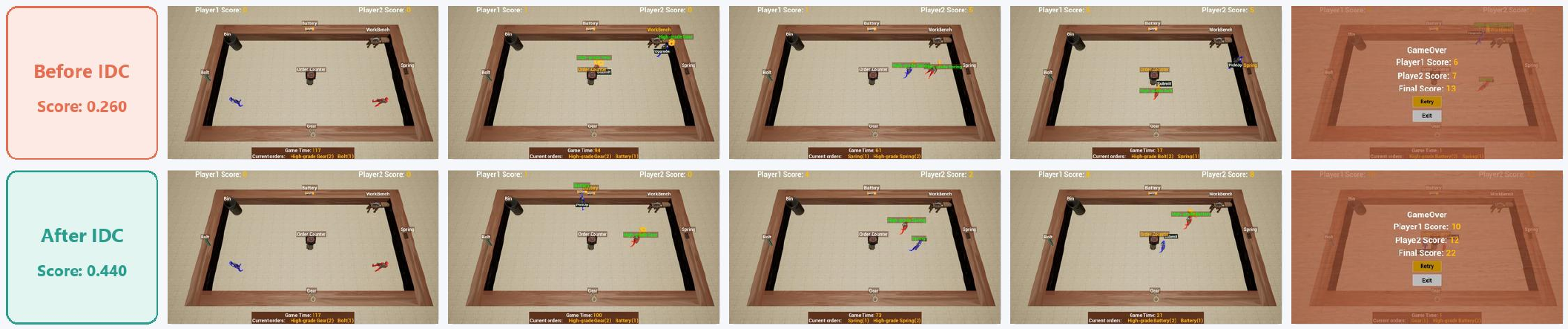}
    \caption{Qualitative comparison on \textit{Shared Floor} using Gemini-3.1-Pro.}
    \label{fig:shared-floor-before-after}
\end{figure*}

\section{Qualitative Comparison with IDC}
IDC skill induction improves the agents' behavior across both environments.
Without IDC, the agents tend to make unstable or inefficient decisions: in the survival-oriented task, the agent fails to consistently maintain a safe position, while in the cooperative task, the agents show weaker coordination and complete fewer objectives.
After IDC, the learned behaviors become more effective and task-aligned.
The agent in the survival task maintains safer positions and achieves a higher score, and the agents in the cooperative task exhibit better coordination and obtain a higher team score. (See Figure~\ref{fig:shared-floor-before-after})

\section{Skill Inspection}
\label{app:idc-skills}

This appendix lists the best measured skill prompt for each game--model pair in the IDC comparison. It explains the divergence discussed in the main text: \textit{LastStand} prompts converge on conservative tile-survival behavior, while \textit{SharedFloor} prompts emphasize cooperative division of labor, station alignment, and order-refresh handling. The scope here is limited to \textit{LastStand} and \textit{SharedFloor}; \textit{ObstacleRun3D} is intentionally excluded.

\section{Visualization}

\begin{figure*}[t]
    \centering
    \includegraphics[width=\textwidth]{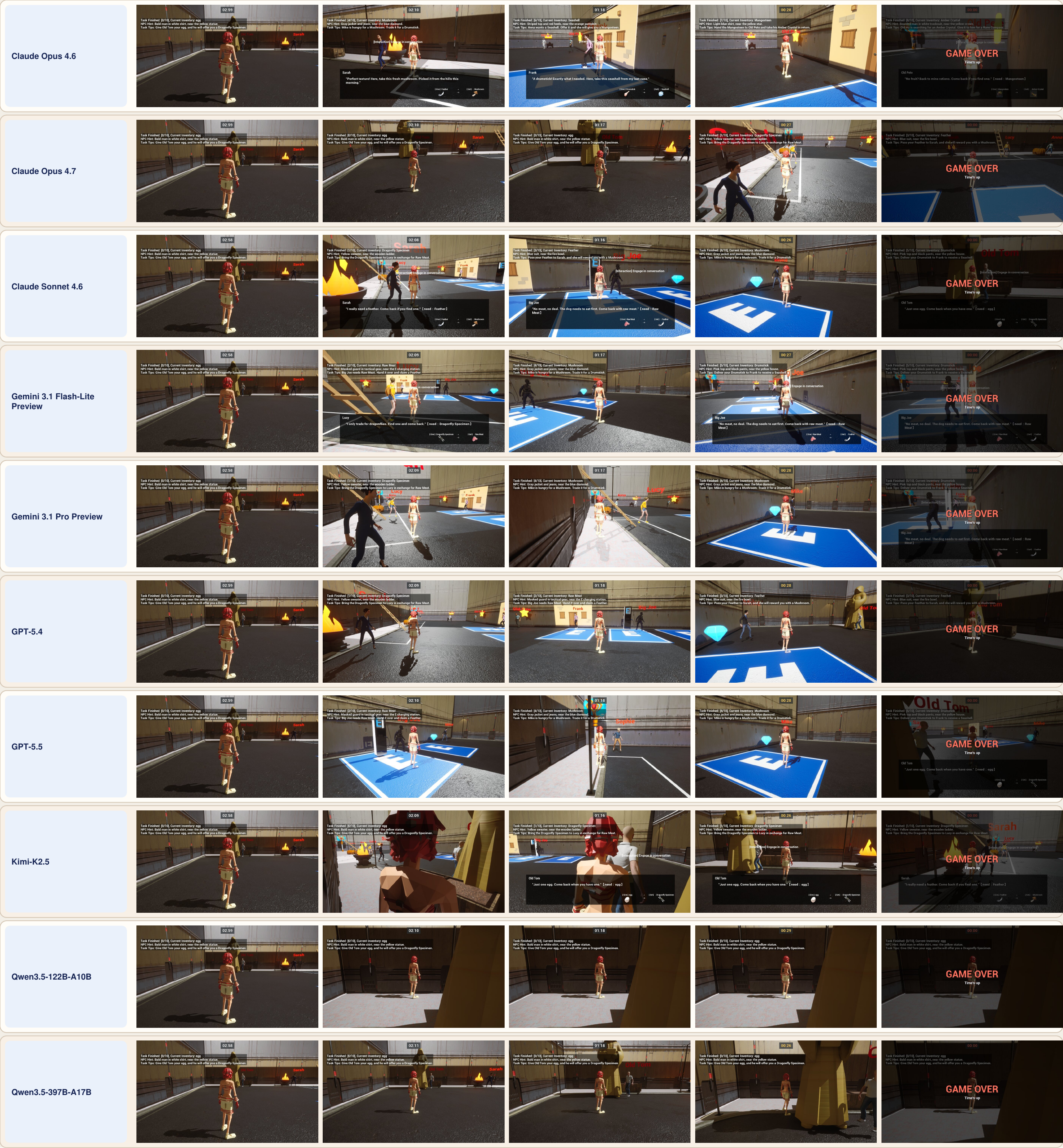}
    \caption{Visualization results for \texttt{cue\_chase}. Each row shows one model, with five sampled frames from the corresponding trajectory.}
    \label{fig:vis-solo-cue-chase}
\end{figure*}

\begin{figure*}[t]
    \centering
    \includegraphics[width=\textwidth]{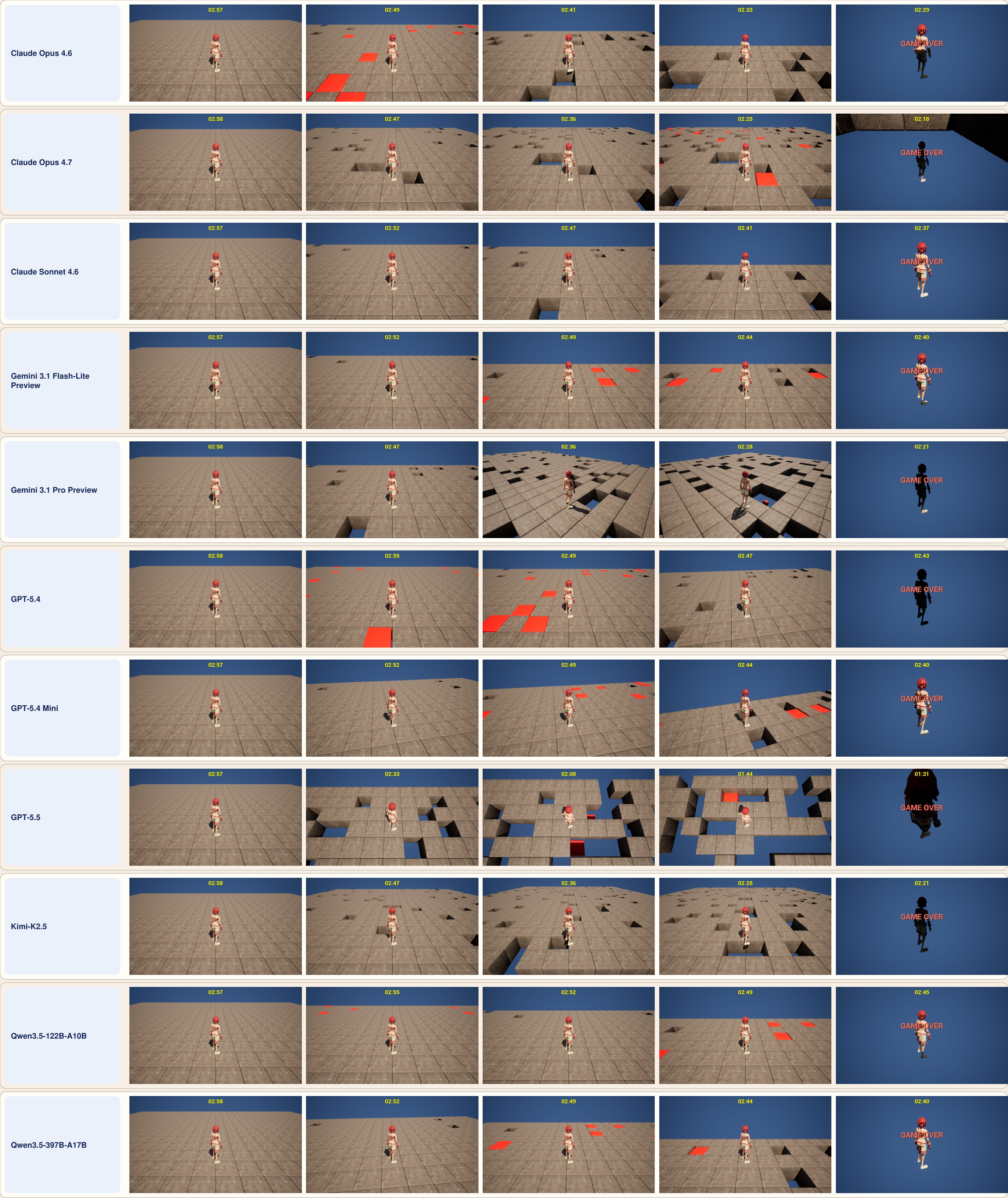}
    \caption{Visualization results for \texttt{last\_stand}. Each row shows one model, with five sampled frames from the corresponding trajectory.}
    \label{fig:vis-solo-last-stand}
\end{figure*}

\begin{figure*}[t]
    \centering
    \includegraphics[width=\textwidth]{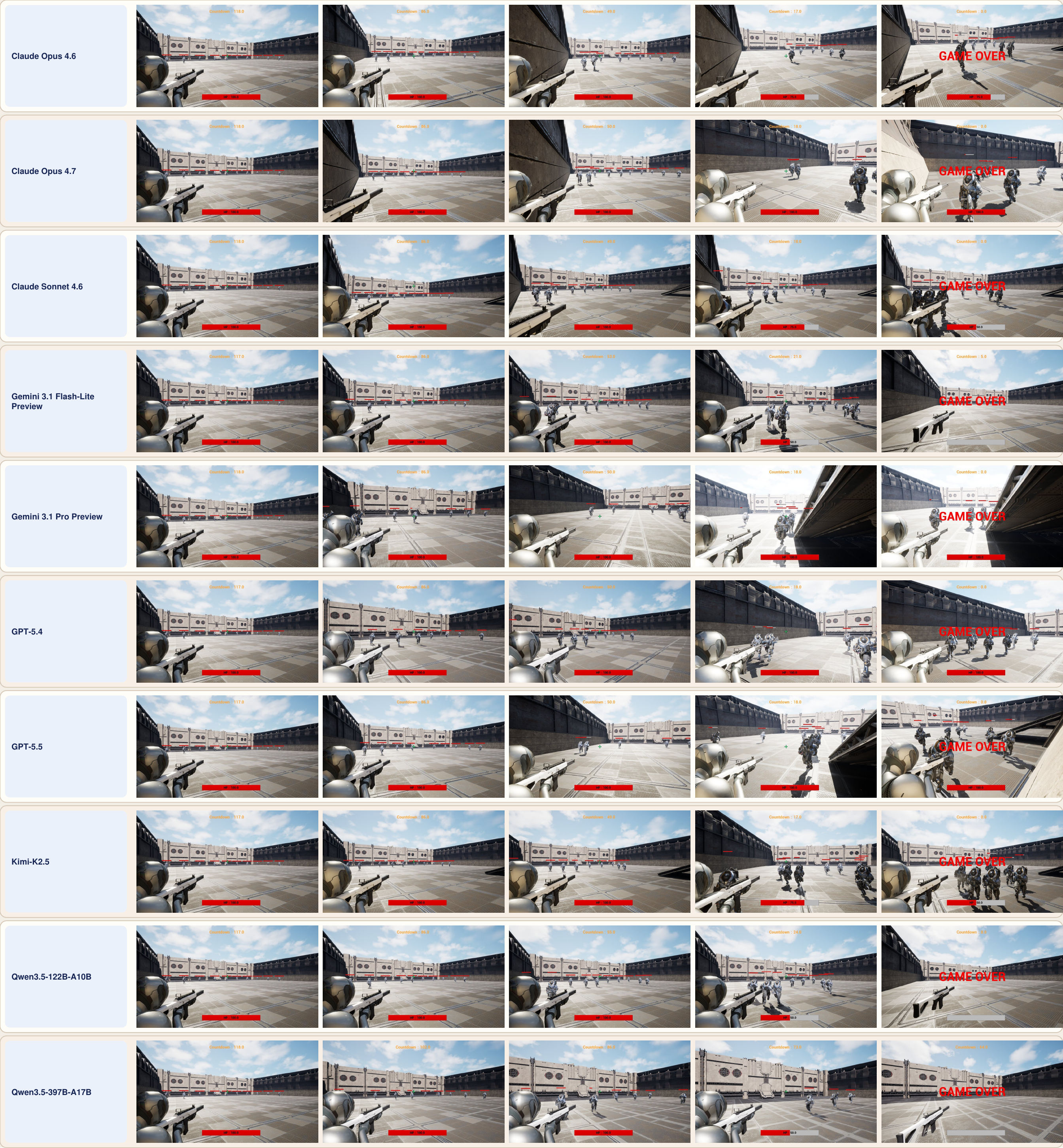}
    \caption{Visualization results for \texttt{monster\_shoot}. Each row shows one model, with five sampled frames from the corresponding trajectory.}
    \label{fig:vis-solo-monster-shoot}
\end{figure*}

\begin{figure*}[t]
    \centering
    \includegraphics[width=\textwidth]{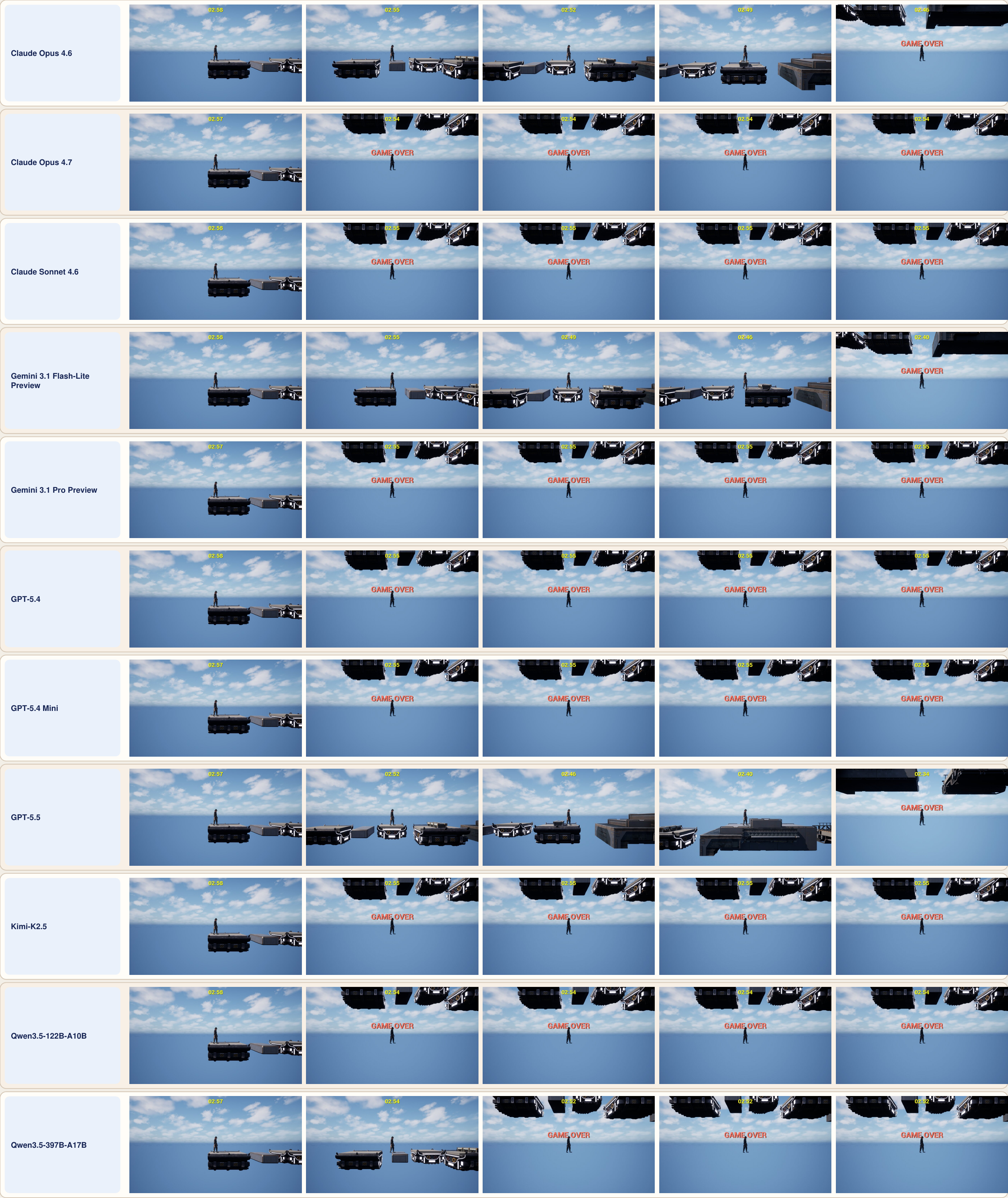}
    \caption{Visualization results for \texttt{obstacle\_run\_2d}. Each row shows one model, with five sampled frames from the corresponding trajectory.}
    \label{fig:vis-solo-obstacle-run-2d}
\end{figure*}

\begin{figure*}[t]
    \centering
    \includegraphics[width=\textwidth]{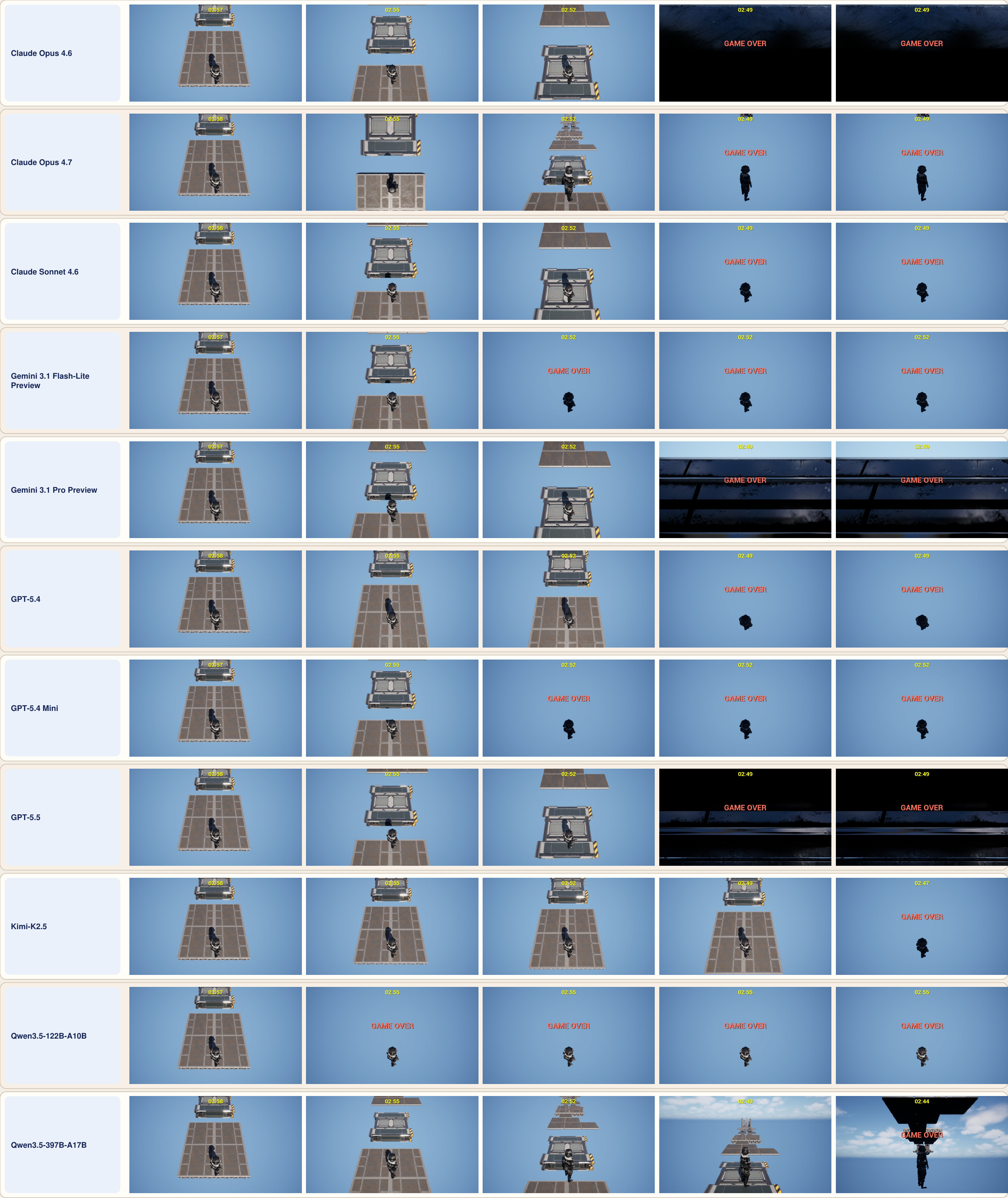}
    \caption{Visualization results for \texttt{obstacle\_run\_3d}. Each row shows one model, with five sampled frames from the corresponding trajectory.}
    \label{fig:vis-solo-obstacle-run-3d}
\end{figure*}

\begin{figure*}[t]
    \centering
    \includegraphics[width=\textwidth]{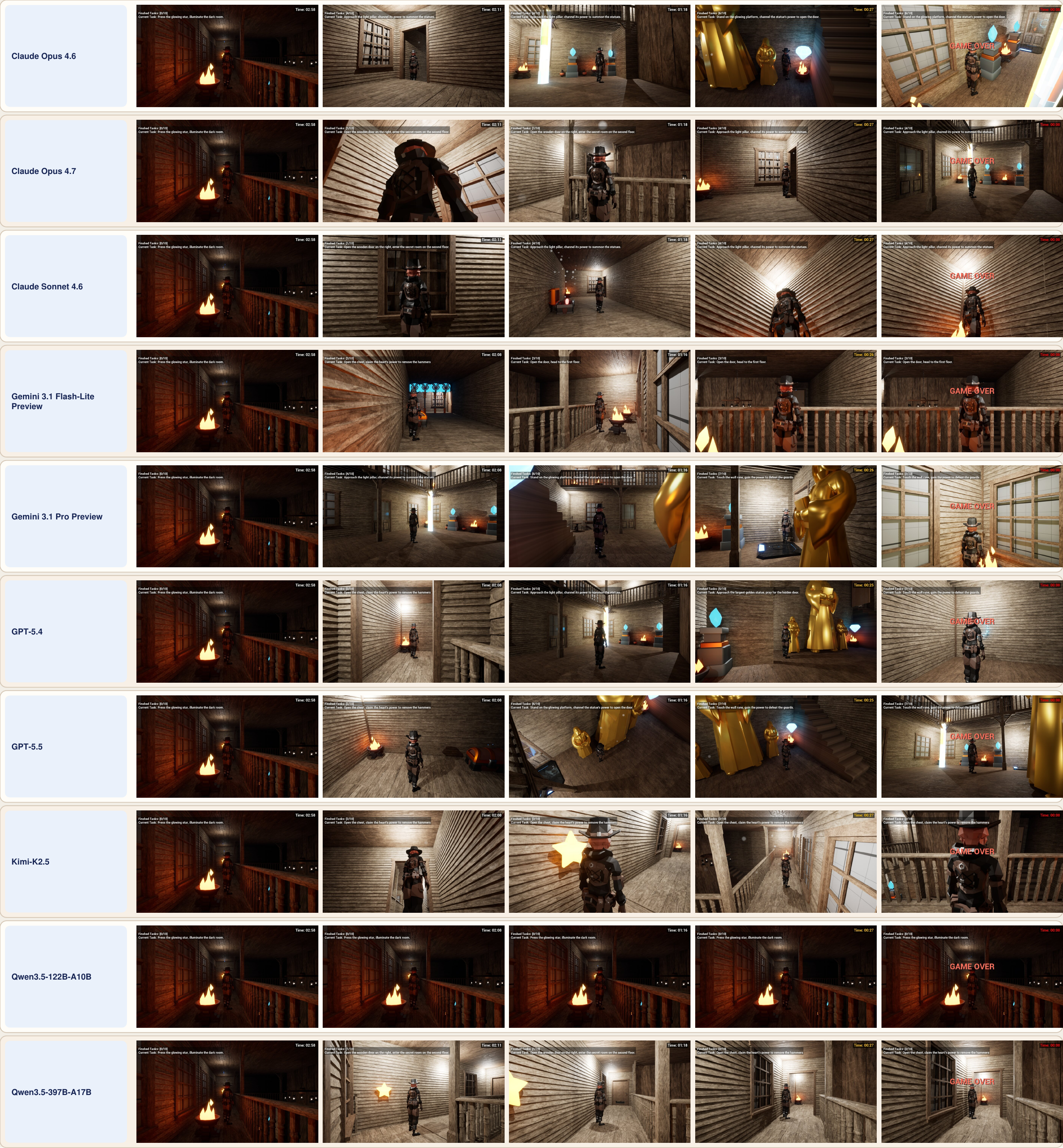}
    \caption{Visualization results for \texttt{scene\_escape}. Each row shows one model, with five sampled frames from the corresponding trajectory.}
    \label{fig:vis-solo-scene-escape}
\end{figure*}

\begin{figure*}[t]
    \centering
    \includegraphics[width=\textwidth]{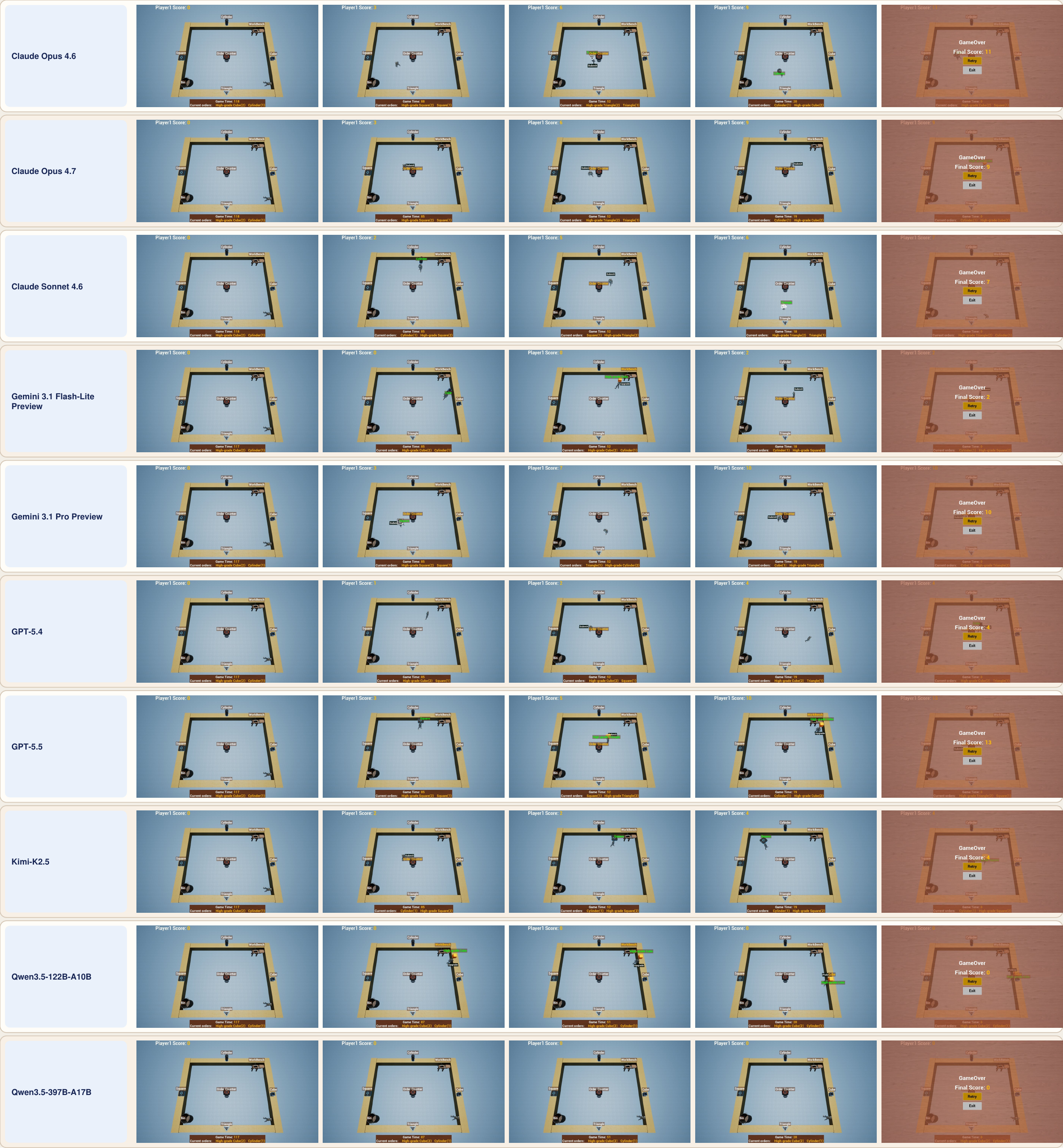}
    \caption{Visualization results for \texttt{solo\_craft}. Each row shows one model, with five sampled frames from the corresponding trajectory.}
    \label{fig:vis-solo-solo-craft}
\end{figure*}

\begin{figure*}[t]
    \centering
    \includegraphics[width=\textwidth]{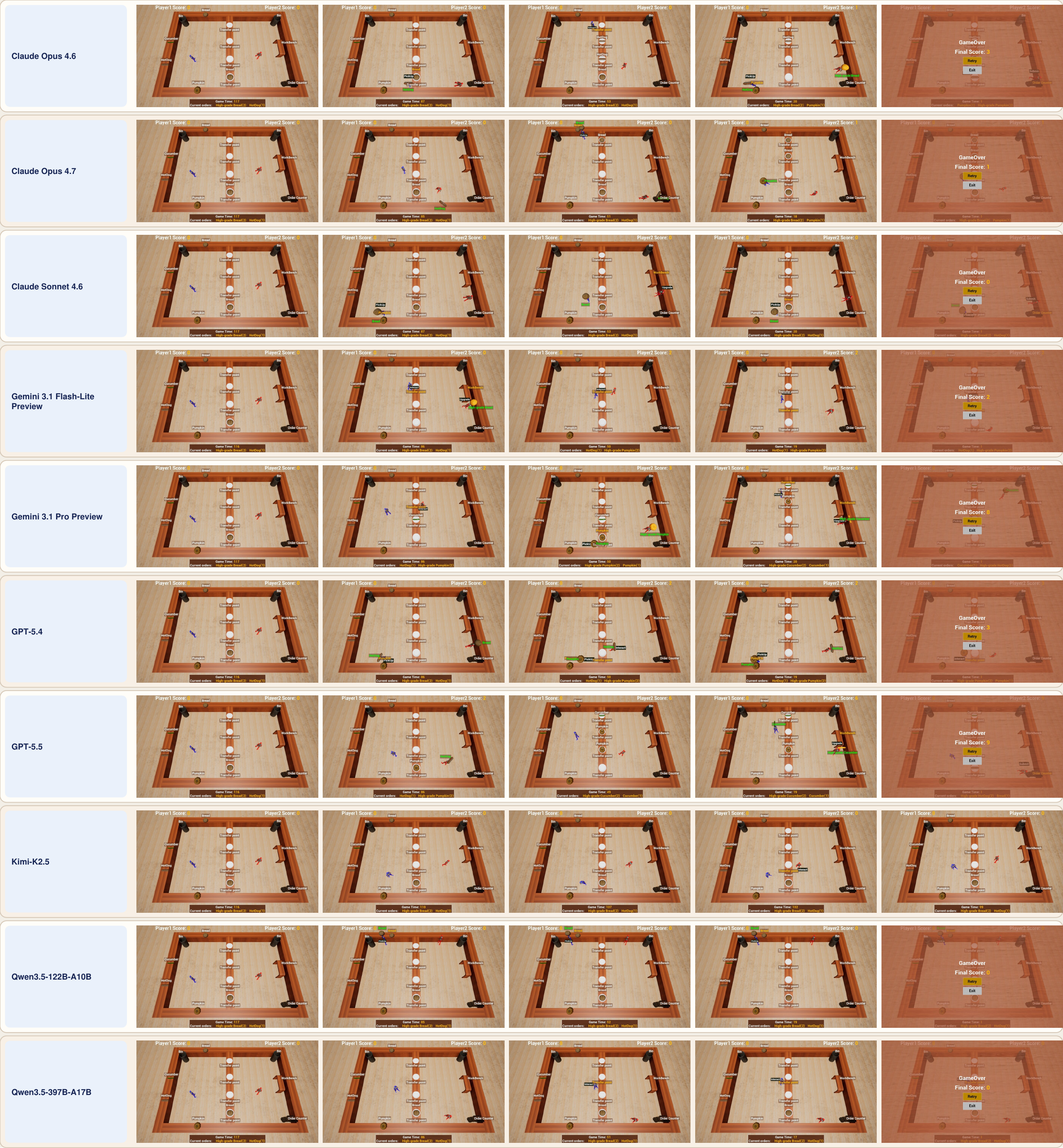}
    \caption{Visualization results for \texttt{handoff\_run}. Each row shows one cooperative model pair, with five sampled frames from the corresponding episode.}
    \label{fig:vis-coop-handoff-run}
\end{figure*}

\begin{figure*}[t]
    \centering
    \includegraphics[width=\textwidth]{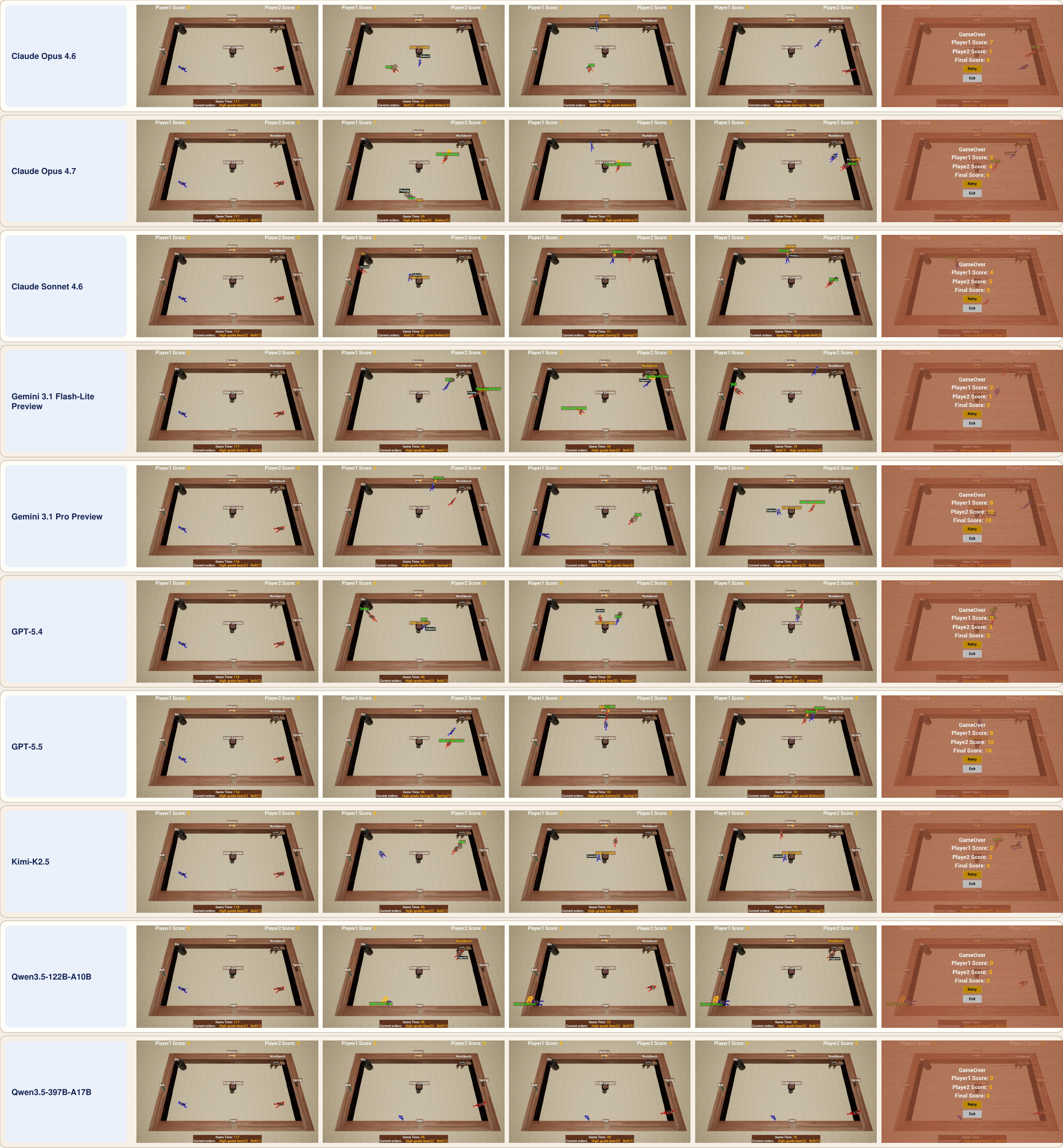}
    \caption{Visualization results for \texttt{shared\_floor}. Each row shows one cooperative model pair, with five sampled frames from the corresponding episode.}
    \label{fig:vis-coop-shared-floor}
\end{figure*}

\begin{figure*}[t]
    \centering
    \includegraphics[width=\textwidth]{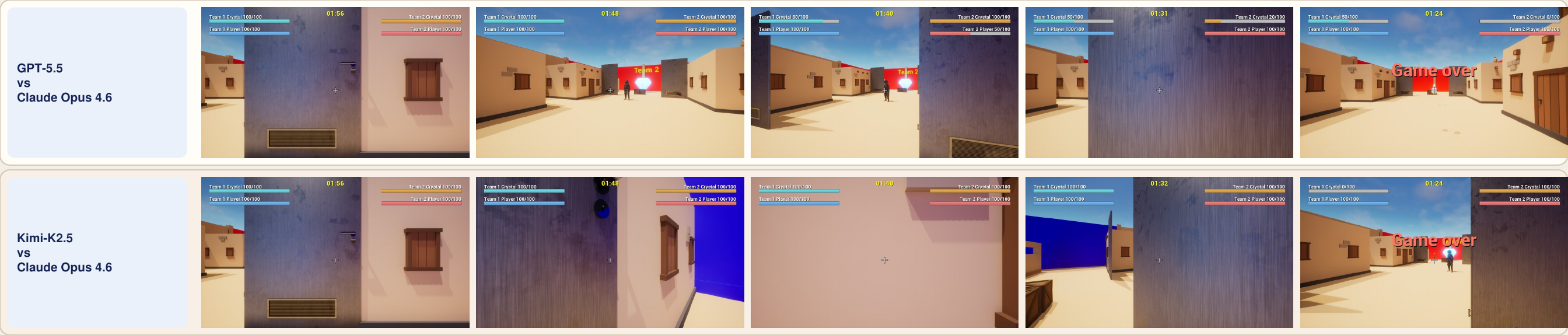}
    \caption{Visualization results for \texttt{crystal\_guard}. Each row shows one representative PvP matchup, with five sampled frames from the corresponding match.}
    \label{fig:vis-pvp-crystal-guard}
\end{figure*}

\begin{figure*}[t]
    \centering
    \includegraphics[width=\textwidth]{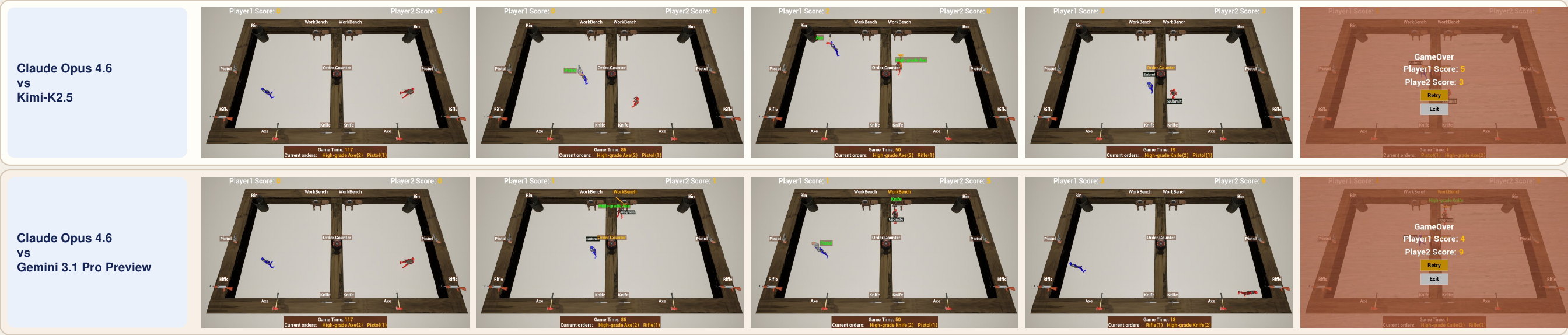}
    \caption{Visualization results for \texttt{midline\_clash}. Each row shows one representative PvP matchup, with five sampled frames from the corresponding match.}
    \label{fig:vis-pvp-midline-clash}
\end{figure*}

\begin{figure*}[t]
    \centering
    \includegraphics[width=\textwidth]{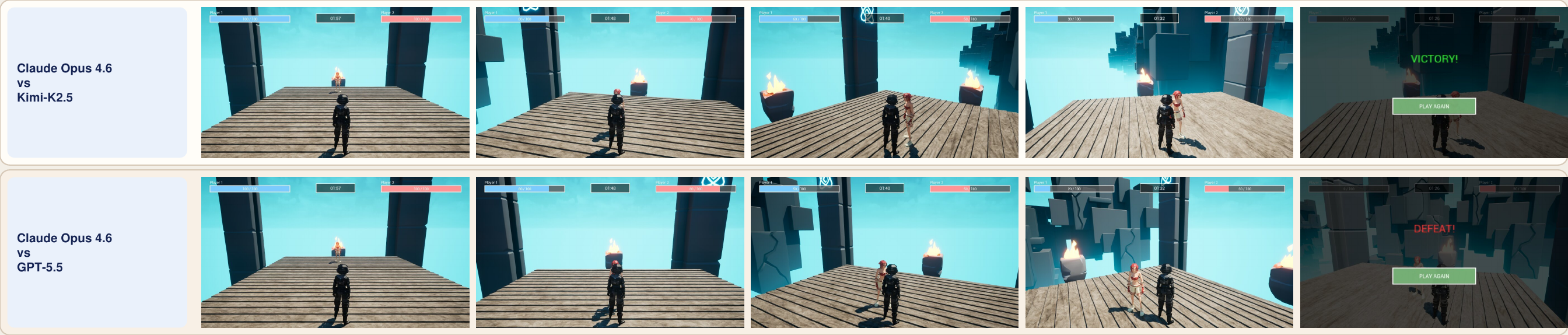}
    \caption{Visualization results for \texttt{sky\_duel}. Each row shows one representative PvP matchup, with five sampled frames from the corresponding match.}
    \label{fig:vis-pvp-sky-duel}
\end{figure*}

Visualization results are shown in Figures~\ref{fig:vis-solo-cue-chase}--\ref{fig:vis-pvp-sky-duel}. 
For each game, we visualize representative trajectories from different models. 
Each row corresponds to one model or matchup, with five sampled frames illustrating the progression of the episode.

\clearpage

\definecolor{SkillInk}{HTML}{334155}
\definecolor{SkillFrame}{HTML}{7A918D}
\definecolor{SkillTitle}{HTML}{6F8581}
\definecolor{SkillBoxBack}{HTML}{F6F8F7}

\newenvironment{skillprompttable}[3]{%
  \def\SkillPromptGame{#1}%
  \def\SkillPromptModel{#2}%
  \begin{table*}[!t]
  \centering
  \small
  \begin{tcolorbox}[
    enhanced,
    width=0.98\textwidth,
    colback=SkillBoxBack,
    colframe=SkillFrame,
    colbacktitle=SkillTitle,
    coltitle=white,
    boxrule=0.75pt,
    arc=2.5mm,
    outer arc=2.5mm,
    left=2.4mm,
    right=2.4mm,
    top=1.8mm,
    bottom=1.8mm,
    title={\textbf{#1}\hfill\texttt{#2}},
    before upper={\color{SkillInk}\setlength{\parindent}{0pt}\setlength{\parskip}{0.35em}}
  ]
}{%
  \end{tcolorbox}
  \caption{Best skill prompt for \SkillPromptGame{} using \texttt{\SkillPromptModel}.}
  \end{table*}
}

\begin{skillprompttable}{LastStand}{claude-opus-4-6}{R7}
This is a survival game: you stand on a tiled platform where tiles progressively turn red (warning) then fall away. Your only goal is to stay alive as long as possible; score increases linearly with survival time.
\par
Your default action is absolute stillness: no movement inputs. Stay still for as many consecutive steps as possible. Every movement risks walking into a gap; stillness on a safe tile carries zero risk. Commit to stillness within the first few steps once you find any safe gray tile.
\par
Camera rotation is completely safe and costs nothing. Use camera turns to scan your surroundings without moving your character. Proactively rotate the camera to check all directions, especially once holes start appearing on multiple sides.
\par
The only reason to move is if the tile directly under your character has turned red. Before moving, explicitly verify: ``Is my tile red?'' If your tile is gray, do not move, regardless of how many red tiles or gaps you see nearby. Adjacent red tiles are not your problem until they become gaps, and even then you only move if your own tile is threatened.
\par
Moving into a gap is always fatal and always worse than staying on a tile that might turn red soon. A red tile gives you at least one full step of warning before it falls. Treat unnecessary movement as more dangerous than a nearby red tile.
\par
When you must move, use the absolute minimum directional inputs: one or two total across all sub-steps. Never use three or more movement inputs in a single step. Overshoot into gaps is the primary killer; a tiny step is always safer than a large one.
\par
Stay near the center of the largest connected cluster of safe gray tiles. Once you reach a reasonable central position, stop immediately and commit to staying there indefinitely until your own tile turns red.
\par
The first few steps are critical: red tiles will appear around the platform but most are not near you. Resist the urge to relocate; if your current tile is gray, stay on it.
\par
Never jump. Jumping adds unpredictability and frequently causes fatal overshoot. Accept being on a smaller island rather than risking a jump.
\par
\end{skillprompttable}

\begin{skillprompttable}{LastStand}{claude-opus-4-7}{R7}
The score grows monotonically each tick you remain alive, so time alive equals score. Almost every move risks landing on a red tile or in a gap.
\par
Never chain forward steps. Repeated \texttt{w ; w ; w} over multiple sub-steps is the strongest predictor of falling; each repeat compounds blindness to what just appeared underfoot.
\par
Avoid jumps unless a gap is unavoidable. Space/jump landings are harder to read than walks; only use when no adjacent solid tile exists.
\par
\end{skillprompttable}

\begin{skillprompttable}{LastStand}{gemini-3.1-pro-preview}{R10}
The goal is to survive as long as possible on a platform where tiles turn red and fall away.
\par
If you are standing on a safe, grey tile, do not move. Output empty actions to stay in place and survive.
\par
Only move when the tile you are standing on turns red.
\par
When your tile turns red, you must visually identify a solid grey block to move to. Empty spaces showing the blue sky below are holes; moving into them will cause you to fall and die.
\par
Look carefully at the adjacent spaces: forward, backward, left, right. Only move in a direction if you clearly see a solid grey tile there. Do not guess or move blindly.
\par
Use minimal steps to reach the adjacent safe tile. Do not use long sequences of movement as you will overshoot and fall into a hole.
\par
\end{skillprompttable}

\begin{skillprompttable}{LastStand}{gpt-5.5}{R1}
Your score mainly improves by staying alive, so prefer stable survival over exploring distant floor.
\par
Keep the view tilted enough toward the floor to judge which tiles are intact, warning-colored, or already gone.
\par
If your current tile turns warning-colored, leave promptly for the nearest connected intact tile; if warning tiles are only nearby, first verify whether standing still is still safe.
\par
Favor short repositioning pulses followed by a pause to reassess, because long continuous movement across fragmented floor caused many early deaths.
\par
Aim to finish each move near the center of the largest connected intact patch, not on edges, seams, narrow bridges, or corners beside voids.
\par
When already centered on a safe small island, hold position until a direct warning appears rather than wandering for marginally larger space.
\par
Treat narrow bridges and isolated transfers as high risk: cross only when the current area is becoming unsafe, then stop and recenter on the wider patch.
\par
Avoid jumping unless there is no connected walking escape from a collapsing or isolated tile.
\par
\end{skillprompttable}

\begin{skillprompttable}{SharedFloor}{claude-opus-4-6}{R8}
This is a cooperative factory game. You and your teammate share the same arena with parts to pick up, a WorkBench to upgrade them, an Order Counter to submit them, and a Bin to discard unwanted items. The team score is the sum of both players' individual scores.
\par
Divide labor immediately: at the start, one player should handle the item closest to them while the other targets a different item. If you see your teammate already moving toward or carrying an item, choose a different one.
\par
Avoid duplication above all else. Before upgrading any item, check if your teammate is carrying or upgrading the same item type. If you cannot confirm they are doing something different, do the basic 1-point version of that order instead. Getting stranded with an unmatched High-grade item wastes 7--10 steps discarding and is the single biggest score killer.
\par
When both a basic order and a High-grade order exist for the same item type, split them: one player does basic, one does High-grade. If you are unsure which your teammate chose, default to basic; it is faster and risk-free.
\par
Only upgrade when you are confident your teammate is not upgrading the same type and at least 20 seconds remain. The upgrade cycle takes 8--10 steps; a failed or duplicated upgrade wastes the entire investment.
\par
Never pick up an item that does not match any current order. Check the orders display before grabbing anything. A High-grade item cannot fulfill a basic order of the same type.
\par
When submitting at the Order Counter, you must be precisely positioned for the ``Submit'' prompt to work. If pressing interact does not register, nudge your position in a distinctly different direction and retry; do not spam interact from the same spot repeatedly.
\par
After upgrading at the WorkBench, verify the label above your character says ``High-grade'' before leaving. If it still shows the basic item name, interact again.
\par
With fewer than about 15 seconds remaining, only attempt basic 1-point orders that do not require upgrading.
\par
After submitting an order, immediately check what new orders appeared and plan your next target before moving.
\par
\end{skillprompttable}

\begin{skillprompttable}{SharedFloor}{claude-opus-4-7}{R3}
This is a 2-player coop kitchen-style game with a shared floor. The bottom HUD lists exactly two open orders at any time: one usually labeled \texttt{High-grade (2)}, the other a base item \texttt{(1)}. High-grade variants must be upgraded at the WorkBench before submission and pay double. Orders rotate the moment a successful submit lands, not on a timer. The round is hard time-limited. Stations include WorkBench, Order Counter, and a Bin for discards.
\par
Use these as heuristics; do not follow them as a script.
\par
Open by proximity, then commit. At step 0, look at spawns and pick whichever order's part you are closer to; the partner takes the other. Do not reverse that assignment in the first several steps because of a guess that the partner might also be going there; abandoning costs more steps than the conflict would. Only switch if you actually see the partner already at that spawn with a PickUp prompt or visibly carrying that item.
\par
Trust the on-screen prompt as ground truth. A visible \texttt{PickUp} / \texttt{Upgrade} / \texttt{Submit} / \texttt{Bin} prompt authoritatively tells you both that you are in range and whether you are or are not carrying the relevant item. The floating green name above your character is unreliable; the prompt is not.
\par
Prompt visible $\rightarrow$ interact, do not reposition. When an interact prompt is on screen, prioritize the interact that frame over more movement; extra movement that frame can carry you back out of range.
\par
Adjacency, not approach. Pickup / Submit / Upgrade triggers on a specific tile, not on raw proximity. If you keep advancing toward a target and no prompt appears, you are misaligned on the perpendicular axis; slide sideways rather than pushing further forward.
\par
Score-credit can lag a step or two behind a successful submit. If you just pressed interact at the counter and the \texttt{team\_score} has not ticked yet, do not mash interact more times in place; the credit usually arrives next observation. If it still has not after a couple of steps, you were misaligned: slide sideways and re-approach.
\par
Re-read the order list every time the score ticks. After any successful submit, yours or the partner's, the orders rotate. Before committing your next move, look at the HUD orders again; if what you are carrying or heading for no longer appears there, redirect immediately.
\par
Bin off dead weight. If orders rotated while you were carrying and what you hold matches neither current order, dump at the Bin and re-grab; do not tour the map with a useless item.
\par
Single re-read on carry-state flip. If your stated inventory has flipped between consecutive frames, pause movement briefly and resolve from the prompt: \texttt{PickUp} showing means you are empty; \texttt{Upgrade} or \texttt{Submit} showing means you are carrying.
\par
Specialization beats symmetry. One player owning the full High-grade loop, pickup $\rightarrow$ WorkBench $\rightarrow$ Submit, while the partner runs the base-item loop is a known high-score pattern, not a coordination failure.
\par
Diversify at the workbench. Do not both queue the same part type at the WorkBench; while one upgrades, the other should be staging the other order's part or running a base submit.
\par
No idle near the counter. Empty-handed after a submit, immediately pick the next part the current orders need; idle frames are pure loss.
\par
Endgame triage. When the round timer is visibly near zero and you are holding a base item, go straight to the Order Counter; do not detour to the WorkBench.
\par
\end{skillprompttable}

\begin{skillprompttable}{SharedFloor}{gemini-3.1-pro-preview}{R9}
Strict division of labor: never work on the same order as your teammate. Check the ``Current orders'' list and observe your teammate's held item and movement direction. If they are moving toward a dispenser, assume they are claiming that item's order and choose a different one.
\par
Preventing the polite dance: if you and your teammate accidentally target the same dispenser, the player who is physically closer should claim it. If you decide to yield, you must commit to a different order and never switch back, even if your teammate also yielded.
\par
Inferring High-grade intent: High-grade orders require taking a base item to the WorkBench to upgrade it. If your teammate is carrying a base item and moving toward the WorkBench, they are claiming the High-grade order for that item.
\par
Verify held item before upgrading: you must physically pick up a base item from its dispenser before you can upgrade it. Do not go to the WorkBench and press interact if your character is empty-handed.
\par
Discarding obsolete items: you can only hold one item at a time. If you are holding an item that is no longer needed, immediately go to the Bin to discard it so you can pick up a new item.
\par
Navigating the Order Counter: the Order Counter is a solid obstacle. If you are stuck walking into it, use perpendicular movement to navigate around it before continuing to your destination.
\par
Combine movement and interaction: do not stop completely to interact with dispensers, the WorkBench, or the Order Counter. Instead, include the interact action in the same step as your final approach movement to save time and ensure you are close enough to trigger the interaction.
\par
\end{skillprompttable}

\begin{skillprompttable}{SharedFloor}{gpt-5.5}{R8}
At each decision, read the active orders and choose the closest unfinished order you can complete now, not the item you were already planning around.
\par
Split labor: when your teammate is clearly carrying or working an order, take a different visible order, a normal cash-in, or a nonblocking support route.
\par
Start an upgrade only when you are visibly holding the matching raw item and can picture the whole workbench-to-counter return; otherwise cash a normal order.
\par
At any station, use one deliberate interaction, then check for a changed held item, score, prompt, or teammate score before repeating.
\par
If an interaction fails, adjust position or reassess whether you are holding the right item instead of standing still or pressing again from the same spot.
\par
After your teammate scores, immediately refresh the order list and treat your carried duplicate as suspect; repurpose it nearby or drop it rather than taking a long disposal trip.
\par
Once you carry a matching finished item, deliver it at the counter, clear the interaction area, then refresh orders and abandon duplicates if your teammate scores first.
\par
Near the end, score only items already held near the counter or already returning from an upgrade; ignore fresh pickups unless they are essentially a submission already.
\par
\end{skillprompttable}

\clearpage

\end{document}